\documentclass[preprint,11pt]{elsarticle}
\usepackage{fullpage}

\makeatletter
\providecommand{\doi}[1]{%
  \begingroup
    \let\bibinfo\@secondoftwo
    \urlstyle{rm}%
    \href{http://dx.doi.org/#1}{%
      doi:\discretionary{}{}{}%
      \nolinkurl{#1}%
    }%
  \endgroup
}
\makeatother

\usepackage{graphicx}
\usepackage[ngerman,english]{babel}
\usepackage{amsxtra, amsfonts, amssymb, amstext}
\usepackage{amsthm}
\usepackage{booktabs}
\usepackage{bbm,dsfont}
\usepackage{amsmath}
\usepackage{nicefrac}
\usepackage{xspace}
\usepackage{url}
\usepackage{graphics,color}
\usepackage[algo2e,ruled,vlined,linesnumbered]{algorithm2e}
\usepackage{multirow}
\newcommand{\assign}{\leftarrow}

\usepackage{hyperref}
\usepackage[hyperref]{xcolor}

\hypersetup{
  colorlinks=true,
}

\hypersetup{
  linkcolor=red,
  urlcolor=red,
  citecolor=red
}

\usepackage{array}
\usepackage{booktabs}
\usepackage{bigdelim}    
\usepackage{graphicx}
\usepackage{wrapfig}
\usepackage{balance} 
\usepackage{epsfig}
\usepackage{subcaption}

\newcommand{\tool}{\textsc{{IOHprofiler}}\xspace}

\newcommand{\oea}{$(1 + 1)$~EA\xspace}

\DeclareMathOperator{\mut}{flip}

\newcommand{\om}{\textsc{OneMax}\xspace}
\newcommand{\onemax}{\om}
\newcommand{\lo}{\textsc{LeadingOnes}\xspace}
\newcommand{\leadingones}{\lo}

\newcommand{\R}{\ensuremath{\mathbb{R}}}

\newcommand{\N}{\ensuremath{\mathbb{N}}}

\newcommand{\Mod}[1]{\ (\mathrm{mod}\ #1)}

\DeclareMathOperator{\Bin}{Bin}

\newcommand{\file}[1]{\texttt{#1}}

\DeclareMathOperator{\id}{id}

\DeclareMathOperator{\cross}{cross}
\DeclareMathOperator{\norm}{norm.}
\DeclareMathOperator{\vars}{var.}
\DeclareMathOperator{\lognormal}{log-n.}
\newcommand{\OM}{\textsc{OM}}

\newcommand{\LinHarm}{\textsc{Harmonic}}
\newcommand{\LO}{\textsc{LO}}
\newcommand{\BV}{\textsc{BV}}

\usepackage{framed,mdframed}
\usepackage{multirow}
\usepackage{array}
\usepackage{bigdelim}
\usepackage{dsfont}

\newcommand{\ga}{$(1 + (\lambda,\lambda))$~GA\xspace}

\allowdisplaybreaks

\journal{Applied Soft Computing}
\begin{document}
\begin{frontmatter}
\title{Benchmarking Discrete Optimization Heuristics with IOHprofiler}
\author[SU]{Carola~Doerr\corref{cor1}}
\ead{Carola.Doerr@lip6.fr}
\author[LU]{Furong~Ye}
\author[MI]{Naama~Horesh}
\author[LU]{Hao~Wang}
\author[MI,THC]{Ofer M.~Shir}
\author[LU]{Thomas~B{\"a}ck}
\cortext[cor1]{Principal corresponding author}
\address[SU]{Sorbonne Universit\'e, CNRS, LIP6, Paris, France}
\address[LU]{Leiden Institute for Advanced Computer Science, Leiden, The Netherlands}
\address[MI]{Migal - The Galilee Research Institute, Upper Galilee, Israel}
\address[THC]{Computer Science Department, Tel-Hai College, Upper Galilee, Israel}

\begin{abstract}
Automated benchmarking environments aim to support researchers in understanding how different algorithms perform on different types of optimization problems. 
Such comparisons provide insights into the strengths and weaknesses of different approaches, which can be leveraged into designing new algorithms and into the automation of algorithm selection and configuration.
With the ultimate goal to create a meaningful benchmark set for iterative optimization heuristics, we have recently released IOHprofiler, a software built to create detailed performance comparisons between iterative optimization heuristics. 

With this present work we demonstrate that IOHprofiler provides a suitable environment for automated benchmarking. We compile and assess a selection of 23 discrete optimization problems that subscribe to different types of fitness landscapes. For each selected problem we compare performances of twelve different heuristics, which are as of now available as baseline algorithms in IOHprofiler. 

We also provide a new module for IOHprofiler which extents the fixed-target and fixed-budget results for the individual problems by ECDF results, which allows one to derive aggregated performance statistics for groups of problems. 
\end{abstract}

\end{frontmatter}

\sloppy{
\section{Introduction}
\label{sec:intro}
Benchmarking optimization solvers aims at supporting practitioners in choosing the best algorithmic technique and its optimal configuration for a given problem.
It is achieved through a systematic empirical assessment and comparison amongst competing techniques on a set of carefully selected optimization problems. Benchmarking may also benefit theoreticians by enhancing mathematically-derived ideas into techniques being broadly applicable in practical optimization. It also constitutes a catalyst in formulating new research questions.
More recently, carefully chosen benchmark problems covering various of the multifaceted characteristics of real-world optimization challenges are also needed as training sets for the automation of algorithm configuration and selection~\cite{kerschke2018survey}.  

The fact that there already exists a broad range of available benchmarking environments demonstrates that the performance assessment of optimization solvers over a set of representative test-problems serves several complementary purposes. 
The nature of the Application Programming Interface, or the identity of the benchmark problems, define together the implementation, and are usually rooted in the sought target(s).  
In the context of discrete optimization, several attempts to construct widely accepted benchmarking environments have been undertaken, 
but these (1) are typically restricted to certain problem classes (often classical NP-hard problems such as SAT, TSP, etc.), (2)  strongly focus on constructive heuristics, which are assumed to have access to the instance data (in contrast to black-box optimization heuristics, which implicitly learn about the problem instance only through the evaluation of potential solutions), or (3) aim to bundle efforts on solving specific real-world problem instances, without the attempt to generate a set of scalable or otherwise generalizable optimization problems. 
Benchmark competitions and crowd-sourcing platforms such as~\cite{WasikABLS16} fall into this latter category.

The few attempts to create a sound benchmarking platform for discrete black-box optimization heuristics, e.g., Weise's optimization benchmarking platform~\cite{weisebench}, have not yet received significant attention from the scientific community. In December 2018, Facebook announced its own benchmarking environment for black-box optimization~\cite{nevergrad}. While their focus is mostly on noisy continuous optimization, the platform also comprises a few discrete problems. 

Interestingly, the situation significantly differs in continuous optimization, where the BBOB workshop series~\cite{BBOB_GECCO2010} and its software framework COCO~\cite{COCO16platform} constitutes a well-established and widely recognized platform for benchmarking derivative-free black-box optimization heuristics. The COCO framework is under constant development. Apart from a noisy test bed, it has in recent years been extended multi-objective~\cite{TusarBHA16} and mixed-integer~\cite{TusarBH19} problems. 
While COCO has been designed to analyze iterative optimization heuristics (IOHs) on different types of problems, its designers have chosen to pre-select these problems that the user can test his/her algorithms on. Benchmarking new problems with COCO requires substantial knowledge of its software design, and is therefore quite time-consuming. 

For the design of \tool~\cite{IOHprofiler}, we have chosen a different way. Our goal is to make the software as flexible as possible, so that the user can easily test his/her algorithms on the problems and with respect to performance criteria of his/her choice, see Section~\ref{sec:iohprofiler} for a brief discussion. The original framework, however, only provided the experimental setup, but did not fix any benchmark problems nor reference algorithms. 

With this work, we contribute to the development of \tool by compiling and evaluating a set of 23 functions for a possible inclusion to a reference set of benchmark problems. 
We also contribute a set of twelve different heuristics that can serve as a first baseline for the performance evaluation of user-defined heuristics. All problems and algorithms have been implemented and integrated in the environment of \tool, so that they are easily accessible for future comparative studies. All performance data is available in our data repository, and can be straightforwardly assessed through the web-based version of \textsc{IOHanalyzer} at \url{http://iohprofiler.liacs.nl/}. 
An important by-product of our contribution is the identification of additional statistics, which should be included within the \tool environment. 
In this respect we contribute a new module for \tool, which can aggregate performance data across different benchmark problems. More precisely, our extension allows to compute ECDF curves for sets of benchmark problems, which complements the previously available statistics in \tool for the assessment of individual benchmark problems.  

This report is an extension of~\cite{IOHprofilerGECCO19}, which has been presented at the 2019 ACM GECCO workshop on Black Box Discrete Optimization Benchmarking (BB-DOB). 

\section{The IOHprofiler Environment}
\label{sec:iohprofiler}

\tool is a new benchmarking environment for detailed, highly modular performance-analysis of iterative optimization heuristics. 
Given algorithms and test problems implemented in C++, Python, or R, \tool outputs a statistical evaluation of the algorithms' performances in the form of the distribution of the fixed-target running time and the fixed-budget function values. 
In addition, \tool also permits tracking the evolution of the algorithms' parameters, which is an attractive feature for the analysis, comparison, and design of (self-)adaptive algorithms. The user selects which information (on top of performance data) is tracked per each algorithm and controls the granularity of the generated records. A documentation of \tool is available at~\cite{IOHprofiler} and at \url{https://iohprofiler.github.io/IOHanalyzer(Post-Processing)/GraphicUserInterface/}.\footnote{The arXiv paper is currently under revision and will be replaced by an updated version soon. Note that in particular the structure of \textsc{IOHexperimenter} has evolved significantly -- this module is not any more based on the COCO framework.} 

\tool consists of two components: \textsc{IOHexperimenter}, a module for processing the actual experiments and generating the performance data, and \textsc{IOHanalyzer}, a post-processing module for compiling detailed statistical evaluations. Data repositories for algorithms, benchmark problems, and performance data are currently under construction. For the time being, the algorithms used for this work as well as the here-described benchmark problems are available in the GitHub repository of \tool, which is available at~\cite{IOhprofiler-github}. Performance data can be assessed through a web-based interface at \url{http://iohprofiler.liacs.nl/}. 

\textsc{IOHexperimenter} is 
designed to handle discrete optimization problems and to facilitate a user-defined selection of benchmark problems. The first version of \textsc{IOHexperimenter} was built upon the COCO framework~\cite{COCO16platform}, but the software has now been restructured to allow for a more flexible selection of benchmark problems -- adding new functions in COCO requires low-level reprogramming of the tool, whereas functions can be added more easily in \tool.

\textsc{IOHanalyzer} has been independently developed from scratch. 
This module can be utilized as a stand-alone tool for the running-time analysis of any algorithm on arbitrary benchmark problems. 
It supports various input file formats, among others the output formats of \textsc{IOHexperimenter} and of COCO. Extensions to other data formats are under investigation.  
\textsc{IOHanalyzer} is designed for a highly interactive evaluation, enabling the user to define their required precision and ranges for the displayed data. The design principle behind \textsc{IOHanalyzer} is a multi-dimensional view on ``performance'', which in our regard comprises at least three objectives: quality, time, and robustness. Unlike other existing benchmark software, \textsc{IOHanalyzer} offers the user to chose which projections of this multi-dimensional data to investigate. In addition to the web-based application at \url{http://iohprofiler.liacs.nl/} and the version on GitHub, \textsc{IOHanalyzer} is also available as CRAN package at \url{https://cran.r-project.org/web/packages/IOHanalyzer/index.html}. 


As mentioned in the introduction, prior to this work \tool only provided the experimental setup for discrete optimization benchmarking, but did not provide a selection of built-in benchmark problems, nor algorithms -- a gap that we start to fill with our present work. Our objectives are two-fold. On the one hand, we want to make a step towards identifying functions that are particularly suitable for discriminating the performance of different IOHs. On the other hand, we also want to demonstrate that \tool can handle large benchmarking projects; for the present work we have performed a total number of 12\,144 experiments: eleven runs of each of the twelve baseline algorithms on a total number of 23 functions in 4 dimensions each. To test the influence of the different instances, an additional 12\,144 runs have been performed.

\section{Suggested Benchmark Functions}
\label{sec:functions}

We evaluate a selection of 23 functions and assess their suitability for inclusion in a standardized discrete optimization benchmarking platform. 
All problems have been implemented and integrated within the \tool framework, and are available at~\cite{IOhprofiler-github}.

Following the discussion in~\cite{ShirDB18}, we restrict our attention to \emph{pseudo-Boolean functions;} i.e., all the suggested benchmark problems are expressed as functions $f:\{0,1\}^n \to \R$. We also pay particular attention to the \emph{scalability} of the problems, with the idea that good benchmark problems should allow to assess performances across different dimensions. 

\paragraph{Conventions} Throughout this work, the variable~$n$ denotes the dimension of the problem that the algorithm operates upon. 
We assume that~$n$ is known to the algorithm; this is a natural assumption, since every algorithm needs to know the decision space that it is requested to search. 
Note though, that the \emph{effective dimension} of a problem can be smaller than~$n$, e.g., due to the usage of ``dummy variables'' that do not contribute to the function values, or due to other reductions of the search space dimensionality (see Section~\ref{sec:W} for examples). 
In practice, we thus only require that~$n$ is an upper bound for the effective number of decision variables.    

For \emph{constrained problems}, such as the N-Queens problem (see Section~\ref{sec:Nqueens}), we follow common practice in the evolutionary computation community and use penalty terms to discount infeasible solutions by the number and magnitude of constraint violations. 

We formulate all problems as \emph{maximization} problems. 
For most, but not for all problems the value of an optimal solution is known. 
Where these maximum function values are not known or are not achieved by any of the algorithms within the allocated computational budget, we evaluate performances with respect to the best solution found by any of the algorithms in any of the runs. 

\paragraph{Notation} 
A search point $x \in \{0,1\}^n$ is written as $(x_1,\ldots,x_n)$. 
By $[k]$ we abbreviate the set $\{1,2,\ldots,k\}$ and by $[0..k]$ the set $[k] \cup \{0\}$. 
All logarithms are to the base 10 and are denoted by $\log$. 
An exception is the natural logarithm, which we denote by $\ln$. 
Finally, we denote by $\id$ the identity function, regardless of the domain. 

\subsection{Rationale Behind The Selection}
\label{sec:mindset}

We briefly discuss the ambition of our work and the requirements that drove the selection of the benchmark problems used for our experiments. 

\paragraph{Ambition} Our ultimate goal is to 
construct a benchmarking suite that covers a wide range of the problem characteristics found in real-world combinatorial  optimization problems. 
A second ambition
lies in building a suitable training set for automated algorithm design, configuration, and selection.

A core assumption of our work is that  
there is no room for a static, ``ultimate'' set of benchmark problems. 
We rather anticipate that a suitable training set should be extendable, to allow users to adjust the selection to their specific needs, but also to reflect advances of the field and to correct misconceptions. 
We particularly foresee a need for augmenting our set of functions, in order to cover landscape characteristics that are not currently present in the problems assessed in this work. 
To put it differently, we present here a first step towards a meaningful collection of discrete optimization benchmarks, but do not claim that that this selection is ``final''. 
In addition, we do not rule out the possibility of removing some of the functions considered below -- for example, if they do not contribute to our understanding of how to distinguish among various heuristics, or if they can be replaced by other problems showing similar effects. 
Indeed, as we will argue in Section~\ref{sec:experiments}, some of the 23 assessed functions do not seem to contribute much to a better discrimination between the tested algorithms, and are therefore evaluated as being obsolete. Note though that this is evaluation crucially depends on the collection of algorithms, so that these functions may have their merit in the assessment of other IOHs.  
As we shall discuss in Section~\ref{sec:outlook}, we are confident that further advances in the research on exploratory landscape analysis~\cite{ExploratoryGECCO2011} could help identify additional problems to be included in the benchmark suite.

\paragraph{Problem Properties} 
As mentioned, we are mostly interested in problems that are arbitrarily scalable with respect to the dimension~$n$. However, as will be the case with the $N$-queens problem, we do not require that there exists a problem instance for \emph{each and every} $n$, but we are willing to accept modest interpretations of scalability. 

For the purposes of our work we demand that evaluating any search point is realizable in reasonable time. 
As a rule of thumb, we are mostly interested in experimental setups that allow one cycle of evaluating all problems within 24 hours for each algorithm. 
In particular, we do \emph{not} address with this work settings that feature \emph{expensive} evaluations. 
We believe that those should be treated separately, as they typically require different solvers than problems allowing for high level of adaption between the evaluations. 

\subsection{Problems vs.~Instances}
\label{sec:unbiasedness}

While we are interested in covering different types of fitness landscapes, we care much less about their actual embedding, and mainly seek to understand algorithms that are invariant under the problem representation. 
In the context of pseudo-Boolean optimization $f:\{0,1\}^n \to \R$, a well-recognized approach to request representation invariance is to demand that an algorithm shows the same or similar performance on any instance mapping each bit string $x \in \{0,1\}^n$ to the function value $f(\sigma(x \oplus z))$, where $z$ is an arbitrary bit string of length $n$, $\oplus$ denotes the bit-wise XOR function, and $\sigma(y)$ is to be read as the string $(y_{\sigma(1)},\ldots,y_{\sigma(n)})$ in which the entries are swapped according to the permutation $\sigma:[n] \to [n]$. 
\tool supports such analysis by allowing to use these transformations (individually or jointly) with randomly chosen $z$ and $\sigma$. 
Using these transformations, we obtain from one particular problem $f$ a whole set of instances $\{ f(\sigma(\cdot \oplus z)) \mid z \in \{0,1\}^n, \sigma \text{ permutation of } [n] \}$, all of which have fitness landscapes that are pairwise isomorphic. 
For further discussions of these \emph{unbiasedness} transformations, the reader is referred to ~\cite{LehreW12,IOHprofiler}.

Apart from unbiasedness, we also focus in this work on \emph{ranking-based heuristics,} i.e., algorithms which only make use of \emph{relative}, and not of \emph{absolute} function values. 
To allow future comparisons with non-ranking-based algorithms, we test all algorithms on instances that are shifted by a multiplicative and an additive offset. 
That is, instead of receiving the values $f(\sigma(x\oplus z))$, only the transformed values $a f(\sigma(x\oplus z)) + b$ are made available to the algorithms. 
We use here again the built-in functionalities of \tool to obtain these transformations.

In the following subsections we describe only the basic instance of each problem, which is identified as instance 1 in \tool. 
We then test all algorithms on instances 1-6 and 51-55, which are obtained from this instance by the transformations described above. 
In theses instances the $\oplus$ and $\sigma$ transformations are separated. 
More precisely, instances 2-6 are obtained from instance 1 by a `$\oplus z$' rotation with a randomly chosen $z \in \{0,1\}^n$, and random fitness offsets $a\in [1/5,5]$, $b \in [-1000,1000]$.  
For instances 51-55 there is no `$\oplus z$' rotation, but the strings are permuted by a randomly chosen $\sigma$ and the ranges for the random fitness offset are chosen as for instances 2-6. 
For each function and each dimension the values of $z$, $\sigma$, $a$, and $b$ are fixed per each instance, but different functions of the same dimensions may have different $z$ and $\sigma$ transformations.

For the reader's convenience we recall that \textsc{IOHexperimenter} uses as pseudo-random number generator the linear congruential generator (LCG), using Schrage's method, the user can find and replace it by other generators in the file \file{IOHprofiler\_random.hpp}. 

\subsection{Overview of Selected Benchmark Problems}
\label{sec:selected}

We summarize here our selected benchmark problems, on which we will elaborate in the subsequent subsections. 
\begin{itemize}
    \item \textbf{F1 and F4-F10:} \onemax and W-model extensions; details in Sections~\ref{sec:OM} and~\ref{sec:W}
    \item \textbf{F2 and F11-F17:} \leadingones; see Sections~\ref{sec:LO} and~\ref{sec:W}
    \item \textbf{F3:} \LinHarm; see Section~\ref{sec:linear}
	\item \textbf{F18:} LABS: Low Autocorrelation Binary Sequences; see Section~\ref{sec:LABS}
	\item \textbf{F19-21:} Ising Models; see Section~\ref{sec:ising}
	\item \textbf{F22:} MIVS: Maximum Independent Vertex Set; see Section~\ref{sec:MIVS}
	\item \textbf{F23:} NQP: N-Queens; see Section~\ref{sec:Nqueens}
\end{itemize}

\subsection{F1: OneMax}
\label{sec:OM}

The \onemax function is the best-studied benchmark problem in the context of discrete evolutionary computation (EC), often referred to as the ``drosophila of EC''. 
It asks to optimize the function 
$$\OM:\{0,1\} \rightarrow [0..n], x \mapsto \sum_{i=1}^n{x_i}.$$ 
The problem has a very smooth and non-deceptive fitness landscape.  
Due to the well-known coupon collector effect (see, for example,~\cite{xxxDubhashiP98} for a detailed explanation of this effect), it is relatively easy to make progress when the function values are small, and the probability to obtain an improving move decreases considerably with increasing function value. 

With the `$\oplus z$' transformations introduced in Section~\ref{sec:unbiasedness}, the \onemax problem becomes the problem of minimizing the Hamming distance to an unknown target string $z \in \{0,1\}^n$. 

\onemax is easily solved in $n$ steps by a greedy hill climber that flips exactly one bit in each iteration, e.g., the one recursively going through the bit string from left to right until no local improvement can be obtained. 
This algorithm is included in our set of twelve reference algorithms as gHC, see Section~\ref{sec:algos}. 
Randomized local search (RLS) and several classic evolutionary algorithms require $\Theta(n \log n)$ function evaluations on \onemax, due to the mentioned coupon collector effect~\cite{GarnierKS99}. 
The \ga from~\cite{DoerrDE15} is the only EA known to optimize \onemax in $o(n \log n)$ time. The self-adjusting version used in our experiments (see Section~\ref{sec:ga}) requires only a linear expected number of function evaluations to locate the optimum~\cite{DoerrD18ga}. 
The best expected running time that any iterative optimization algorithm can achieve is $\Omega(n/\log n)$~\cite{ErdR63}. 
However, it is also known that unary unbiased algorithms cannot achieve average running times better than $\Omega(n \log n)$~\cite{LehreW12,DoerrDY16}. 
A more detailed survey of theoretical running-time results for \onemax can be found in~\cite{DoerrYRWB18}, and a survey of lower bounds (in terms of black-box complexity results) can be found in~\cite{Doerr18BBC}. 
That \onemax is interesting beyond the study of theoretical aspects of evolutionary computation has been argued in~\cite{Thierens09}. We believe that \onemax plays a similar role as the sphere function in continuous domains, and should be added to each benchmark set: it is not very time-consuming to evaluate, and can provide a first basic stress test for new algorithm designs.

\subsection{F2: LeadingOnes}
\label{sec:LO}

Among the non-separable functions, the \leadingones function is certainly the one receiving most attention in the theory of EC community. 
The \leadingones problem asks to maximize the function $$\LO:\{0,1\}^n \to [0..n], x\mapsto \max \{i \in [0..n] \mid \forall j \le i: x_j=1\} = \sum_{i=1}^n{\prod_{j=1}^i{x_j}},$$ which counts the number of initial ones.  

Most EAs require quadratic running time to solve \leadingones, see again~\cite{DoerrYRWB18} or the full version of~\cite{Doerr18domi} for a summary of theoretical results. It is known that all elitist (1+1)-type algorithms~\cite{DoerrL17LO} and all unary unbiased~\cite{LehreW12} are restricted to an $\Omega(n^2)$ expected running time. 
However, some problem-tailored algorithms optimizing \leadingones in sub-quadratic expected optimization time have been designed~\cite{AfshaniADLMW19,DoerrW11EA,DoerrJKLWW11}. It is also known that the best-possible expected running time of any iterative optimization heuristic is $\Theta(n \log\log n)$~\cite{AfshaniADLMW19}. Similar to \onemax, we argue that \leadingones should form a default benchmark problem: it is fast to evaluate and can point at fundamental issues of algorithmic designs, see also the discussions in Section~\ref{sec:experiments}.

%
%
\subsection{F3: A Linear Function with Harmonic Weights}
\label{sec:linear}

Two extreme linear functions are \onemax with its constant weights and binary value $\BV(x)=\sum_{i=1}^{n}{2^{n-i}x_i}$ with its exponentially decreasing weights. An intermediate linear function is 
$$f:\{0,1\}^n \to \R, x \mapsto \sum_{i} i x_i$$ with harmonic weights, which was suggested to be considered in~\cite{ShirDB18}. We add this linear function to our assessment as F3. 

Several results mentioned in the paragraph about \onemax apply more generally to linear functions, and hence to the special case F3. For example, it is well known that the \oea and RLS need $\Theta(n \log n)$ function evaluations, on average, to optimize any linear function~\cite{DrosteJW02}. No unary unbiased black-box algorithm can achieve a better expected running time~\cite{LehreW12}. It is also known that for the \oea the best static mutation rate for optimizing any linear function is $1/n$~\cite{Witt13j}. The (unrestricted) black-box complexity of linear functions, however, is not known. However, we easily see that the mentioned greedy hill climber gHC described in Section~\ref{sec:OM} optimizes F3 in at most $n+1$ queries.

\subsection{F4-F17: The W-model}
\label{sec:W}

In~\cite{Wmodel} a collection of different ways to ``perturb'' existing benchmark problems in order to obtain new functions of scalable difficulties and landscape features has been suggested, the so-called W-model. These W-model transformations can be combined arbitrarily, resulting in a huge set of possible benchmark problems. In addition, these transformations can, in principle, be superposed to any base problem, giving yet another degree of freedom. Note here that the original work~\cite{Wmodel} as well as the existing empirical evaluations~\cite{Wmodelpractice} only consider \onemax as underlying problem, but there is no reason to restrict the model to this function. 
We expect that in the longer term, the W-model, similarly to the well-known NK-landscapes~\cite{NKlandscapes} may constitute an important building block for a scalable set of discrete benchmark problems. More research, however, is needed to understand how the different combinations influence the behavior of state-of-the-art heuristic solvers. In this work, we therefore restrict our attention to instances in which the different components of the W-model are used in an isolated way, see Section~\ref{sec:Wmodelfunctions}.. The assessment of combined transformations clearly forms a promising line for future work.

\subsubsection{The Basic Transformations}
\label{sec:Wbasic}

The W-model comprises four basic transformations, and each of these transformations is parametrized, hence offering a huge set of different problems already. We provide a brief overview of the W-model transformations that are relevant for our work. A more detailed description can be found in the original work~\cite{Wmodel}. For some of the descriptions below we deviate from the exposition in~\cite{Wmodel}, because in contrast to there, we consider \emph{maximization} as objective, not minimization. Note also that we write $x=(x_1,\ldots,x_n)$, whereas in~\cite{Wmodel} the strings are denoted as $(x_{n-1}, x_{n-2}, \ldots, x_1,x_0)$. Note also that the reduction of dummy variables is our own extension of the W-model, not originally proposed in~\cite{Wmodel}.

\begin{enumerate}
	\item \textbf{Reduction of dummy variables $W(m,\ast,\ast,\ast)$:} a reduction mapping each string $(x_1, \ldots, x_n)$ to a substring $(x_{i_1}, \ldots, x_{i_m})$ for randomly chosen, pairwise different $i_1,\ldots, i_m \in [n]$. This modification models a situation in which some decision variables do not have any or have only negligible impact on the fitness values. Thus, effectively, the strings $(x_1,\ldots,x_n)$ that the algorithm operates upon are reduced to substrings $(x_{i_1},\ldots,x_{i_m})$ with $1 \le i_1 < i_2 < \ldots < i_m \le n$. 
	
	We note that such scenarios have been analyzed theoretically, and different ways to deal with this \emph{unknown solution length} have been proposed. Efficient EAs can obtain almost the same performance (in asymptotic terms) than EAs ``knowing'' the problem dimension~\cite{EinarssonLGMMSW18,DoerrDK17}.
	
	Dummy variables are also among the characteristics of the benchmark functions contained in Facebook's nevergrad platform~\cite{nevergrad}, which might be seen as evidence for practical relevance.  
	
	\underline{Example:} With $n=10$, $m=5$, $i_1=1$, $i_2=2$, $i_3=4$, $i_4=7$, $i_5=10$, the bit string $(1010101010)$ is reduced to $(10010)$.
	\item \textbf{Neutrality $W(\ast,\mu,\ast,\ast)$:} The bit string $(x_1,\ldots,x_n)$ is reduced to a string $(y_1,\ldots,y_m)$ with $m=n/\mu$, where $\mu$ is a parameter of the transformation. For each $i \in [m]$ the value of $y_i$ is the majority of the bit values in a size-$\mu$ substring of $x$. More precisely, $y_i=1$ if and only if there are at least $\mu/2$ ones in the substring $(x_{(i-1)\mu+1},x_{(i-1)\mu+2},\ldots,x_{i\mu})$.\footnote{Note that with this formulation there is a bias towards ones in case of a tie. We follow here the suggestion made in~\cite{Wmodel}, but we note that this bias may have a somewhat complex impact on the fitness landscape. For our first benchmark set, we therefore suggest to use this transformation with odd values for $\mu$ only.} When $n/\mu \notin \N$, the last $n-\mu \lfloor n/\mu \rfloor$ remaining bits of $x$ not fitting into any of the blocks are simply deleted; that is, we have $m=\lfloor n/\mu \rfloor$ and the entries $x_i$ with $i> \mu \lfloor n/\mu \rfloor$ do not have any influence on~$y$ (and, thus, no influence on the function value).
	
	\underline{Example:} With $n=10$ and $\mu=3$ the bit string $(1110101110)$ is reduced to $(101)$. 
	\item \textbf{Epistasis $W(\ast,\ast,\nu,\ast)$:} The idea is to introduce local perturbations to the bit strings. To this end, a string $x=(x_1,\ldots,x_n)$ is divided into subsequent blocks of size $\nu$. Using a permutation $e_{\nu}:\{0,1\}^{\nu} \to \{0,1\}^{\nu}$, each substring $(x_{(i-1)\nu+1},\ldots,x_{i\nu})$ is mapped to another string $(y_{(i-1)\nu+1},\ldots,y_{i\nu})=e_{\nu}((x_{(i-1)\nu+1},\ldots,x_{i\nu}))$. The permutation $e_{\nu}$ is chosen in a way that Hamming-1 neighbors $u,v \in \{0,1\}^{\nu}$ are mapped to strings of Hamming distance at least $\nu-1$. Section~2.2 in~\cite{Wmodel} provides a construction for such permutations. For illustration purposes, we repeat below the map for $\nu=4$, which is the parameter used in our experiments. This example can also be found, along with the general construction, in~\cite{Wmodel}. 
	\begin{align*}
    e_4(0000) &= 0000 \quad\quad  e_4(0001) = 1101 \quad\quad  e_4(0010) = 1011 \quad\quad  e_4(0011) = 0110\\
    e_4(0100) &= 0111 \quad\quad  e_4(0101) = 1010 \quad\quad  e_4(0110) = 1100 \quad\quad  e_4(0111) = 0001\\
    e_4(1000) &= 1111 \quad\quad  e_4(1001) = 0010 \quad\quad  e_4(1010) = 0100 \quad\quad  e_4(1011) = 1001\\
    e_4(1100) &= 1000 \quad\quad  e_4(1101) = 0101 \quad\quad  e_4(1110) = 0011 \quad\quad  e_4(1111) = 1110
\end{align*}

When $n/\nu \notin \N$, the last bits of $x$ are treated by $e_{n-\nu \lfloor n/\nu \rfloor}$; that is, the substring $(x_{\nu \lfloor n/\nu \rfloor +1}, x_{\nu \lfloor n/\nu \rfloor +2}, \ldots, x_n)$ is mapped to a new string of the same length via the function $e_{n-\nu \lfloor n/\nu \rfloor}$. 
	
	\underline{Example:} With $n=10$, $\nu=4$, and the permutation $e_{4}$ provided above, the bit string $(1111011101)$ is mapped to $(1110000110)$, because $e_4(1111)=1110$ and $e_4(0111)=0001$ and $e_2(01)=10$. 
	\item \textbf{Fitness perturbation $W(\ast,\ast,\ast,r)$:} With these transformations we can determine the \emph{ruggedness} and \emph{deceptiveness} of a function. Unlike the previous transformations, this perturbation operates on the function values, not on the bit strings. To this end, a \emph{ruggedness} function $r:\{f(x) \mid x \in \{0,1\}^n \}=:V \to V$ is chosen. The new function value of a string $x$ is then set to $r(f(x))$, so that effectively the problem to be solved by the algorithm becomes $r \circ f$. 
	
	To ease the  analysis, it is required in~\cite{Wmodel} that the optimum 
	$v_{\max}=\max \{f(x) \mid x \in \{0,1\}^n \}$ 
	does not change, i.e., $r$ must satisfy that $r(v_{\max}\})= v_{\max}$ and $r(i)<v_{\max}$ for all $i<v_{\max}$. It is furthermore required in~\cite{Wmodel} that the ruggedness functions $r$ are permutations (i.e., one-to-one maps). Both requirements are certainly not necessary, in the sense that additional interesting problems can be obtained by violating these constrains. We note in particular that in order to study \emph{plateaus} of equal function values, one might want to choose functions that map several function values to the same value. We will include one such example in our testbed, see Section~\ref{sec:Wsettings}.
	
	It should be noted that all functions of unitation (i.e., functions for which the function value depends only on the \onemax value of the search point, such as \textsc{Trap} or \textsc{jump}) can be obtained from a superposition of the fitness perturbation onto the \onemax problem. 
	
	\underline{Example:} The well-known, highly deceptive \textsc{Trap} function can be obtained by superposing the permutation $r:[0..n] \to [0..n]$ with $r(i)=n-1-i$ for all $1 \le i \le n$ and $r(n)=n$.  
\end{enumerate}

\subsubsection{Combining the Basic W-model Transformations}
\label{sec:Wcombi}

We note that any of the four W-model transformations can be applied independently of each other. The first three modification can, in addition, be applied in an arbitrary order, with each order resulting in a different benchmark problem. In line with the presentation in~\cite{Wmodel}, we consider in our implementation only those perturbations that follow the order given above. Each set of W-model transformations can be identified by a string $(\{i_1,\ldots,i_m\},\mu,\nu,r)$ with $m\le n$, $1 \le i_1< \ldots <i_m \le n$, $\mu \in [n]$, $\nu \in [n]$, and $r:V \to V$, all to be interpreted as in the descriptions given in Section~\ref{sec:Wbasic} above. Setting $\{i_1,\ldots,i_m\}=[n]$, $\mu=1$, $\nu=1$, and/or $r$ as the identity function on $V$ corresponds to not using the first, second, third, and/or forth transformation, respectively.

As mentioned, the W-model can in principle be superposed on any benchmark problem. The only complication is that the search space on which the algorithm operates and the search space on which the benchmark problem is applied are not the same when $m<n$ or $\mu>1$. More precisely, while the algorithm operates on $\{0,1\}^n$, the base problem has to be a function $f:\{0,1\}^s \to \R$ with $s=\lfloor m/\mu \rfloor$. We call $s$ the \emph{effective dimension} of the problem. 
When $f$ is a scalable function defined for any problem dimension $s$---this is the case for most of our benchmark functions---we just reduce to the $s$-dimensional variant of the problem. When $f$ is a problem that is only defined for a fixed dimension $n$, the algorithms should operate on the search space $\{0,1\}^{\ell}$ with $\ell \ge \mu s$ and $\ell- \mu s$ depending on the reduction that one wishes to achieve by the first transformation, the removal of dummy variables.

\subsubsection{Selected W-Model Transformations}
\label{sec:Wmodelfunctions}\label{sec:Wsettings}


In contrast to existing works cited in~\cite{Wmodel,Wmodelpractice}, we do not only study superpositions of W-model transformations to the \onemax problems (functions F4-F10), but we also consider \leadingones as a base problem (F11-17). This allows us to study the effects of the transformations on a well-understood separable and a well-understood non-separable problem. As mentioned, we only study individual transformations, and not yet combinations thereof. 

\begin{figure}[t]
\centering
\includegraphics[width=\linewidth]{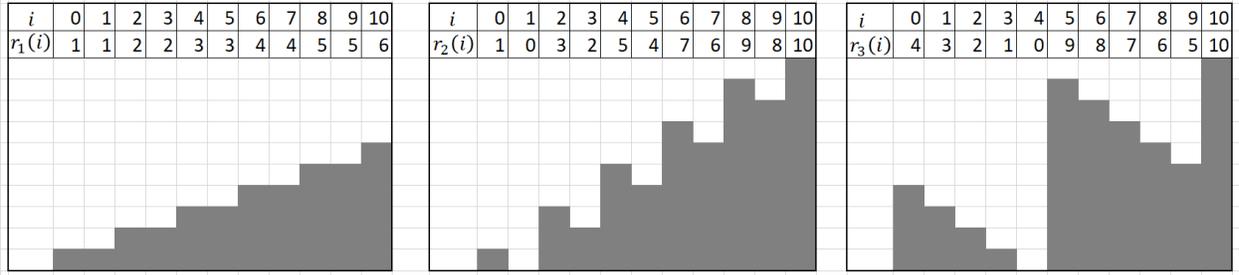}
\caption{The ruggedness functions $r_1$, $r_2$, and $r_3$.\label{app:fig:R1}}
\end{figure}

We consider the reduction of $[n]$ to subsets of size $n/2$ and $0.9n$, i.e., only half and 90\% of the bits, respectively, contribute to the overall fitness. 
We consider neutrality transformations of size $\mu=3$, and we consider the epistasis perturbation of size $4$. Finally, we consider the following ruggedness functions, where we denote by $s$ the size of the effective dimension (see Section~\ref{sec:Wcombi} for a discussion) and recall that both the $s$-dimensional \onemax and \leadingones functions take values in $[0..s]$. These functions are illustrated for $s=10$ in Figure~\ref{app:fig:R1}.
\begin{itemize}
    \item $r_1:[0..s] \to [0..\lceil s/2 \rceil+1]$ with $r_1(s)= \lceil s/2 \rceil +1$ and $r_1(i)=\lfloor i/2 \rfloor+1$ for $i<s$ and even $s$, and $r_1(i)=\lceil i/2 \rceil+1$ for $i<s$ and odd $s$. 
    \item $r_2:[0..s] \to [0..s]$ with $r_2(s)=s$, $r_2(i)=i+1$ for $i \equiv s \Mod{2}$ and $i<s$, and $r_2(i)=\max\{i-1,0\}$ otherwise.
    \item $r_3:[0..s] \to [-5..s]$ with $r_3(s)=s$ and $r_3(s-5j+k)=s-5j+(4-k)$ for all $j \in [s/5]$ and $k\in [0..4]$ and 
    $r_3(k)=s -  (5\lfloor s/5 \rfloor  - 1 )-  k$  for $k \in [0..s - 5\lfloor s/5 \rfloor -1]$. 
\end{itemize}
We see that function $r_1$ keeps the order of the function values, but introduces small plateaus of the same function value. In contrast to $r_1$, function $r_2$ is a permutation of the possible function values. It divides the set of possible non-optimal function values $[0..s-1]$ into blocks of size two (starting at $s-1$ and going in the inverse direction) and interchanges the two values in each block. When $s$ is odd, the value $0$ forms its own block with $r_1(0)=0$. Similarly, $r_3$ divides the set of possible function values in blocks of size 5 (starting at $s-1$ and going in inverse direction), and reverses the order of function values in each block.  

Summarizing all these different setups, the functions F4-F17 are defined as follows: 
\begin{center}
\begin{tabular}{ll}
   F4: $\onemax + W([n/2],1,1,\id)$ \quad & F11: $\leadingones + W([n/2],1,1,\id)$\\
	F5: $\onemax + W([0.9n],1,1,\id)$ & 	F12: $\leadingones + W(0.9n,1,1,\id)$\\
	F6: $\onemax + W([n],\mu=3,1,\id)$ & F13: $\leadingones + W([n],\mu=3,1,\id)$\\
	F7: $\onemax + W([n],1,\nu=4,\id)$ & F14: $\leadingones + W([n],1,\nu=4,\id)$\\
	F8: $\onemax + W([n],1,1,r_1)$ & F15: $\leadingones + W([n],1,1,r_1)$\\
	F9: $\onemax + W([n],1,1,r_2)$ & F16: $\leadingones + W([n],1,1,r_2)$\\
	F10: $\onemax + W([n],1,1,r_3)$ & F17: $\leadingones + W([n],1,1,r_3)$
\end{tabular}
\end{center}

\subsubsection{W-model vs. Unbiasedness Transformations and Fitness Scaling}
\label{sec:Wunbiased}

To avoid confusion, we clarify the sequence of the transformations of the W-model and the unbiasedness and fitness value transformations discussed in Section~\ref{sec:unbiasedness}. Both the re-ordering of the string by the permutation $\sigma$ and the XOR with a fixed string $z \in \{0,1\}^n$ are executed \emph{before} the transformations of the W-model are applied, while the multiplicative and additive scaling of the function values is applied to the result \emph{after} the fitness perturbation of the W-model. 

\textbf{Example:} Assume that the instance is generated from a base problem $f:\{0,1\}^n \to \R$, that the unbiasedness transformations are defined by a permutation $\sigma:[n] \to [n]$ and the string $z \in \{0,1\}^n$, the fitness scaling by a multiplicative scalar $b >0$ and an additive term $a \in \R$. Assume further that the W-model transformations are defined by the vector $({i_1,\ldots,i_m},\mu,\nu,r)$. 
For each queried search point $x \in \{0,1\}^n$, the algorithm receives the function value $af(W(\sigma(x) \oplus z))+b$, 
where $\sigma(x)=(x_{\sigma(1)}, \ldots, x_{\sigma(n)})$ and $W:\{0,1\}^n \to \R$ denotes the function that maps each string to the fitness value defined via the W-transformations $({i_1,\ldots,i_m},\mu,\nu,r)$.

\subsection{F18: Low Autocorrelation Binary Sequences}
\label{sec:LABS}

Obtaining binary sequences possessing a high merit factor, also known as the Low-Autocorrelation Binary Sequence (LABS) problem, constitutes a grand combinatorial challenge with practical applications in radar engineering and measurements \cite{ShapiroPRL1968,Pasha2000}. It also carries several open questions concerning its mathematical nature.
Given a sequence of length $n$, $S=\left(s_1,\ldots,s_n\right)$ with 
$s_i \in \{-1, +1\}$,
the merit factor is proportional to the reciprocal of the sequence's autocorrelation. 
The LABS optimization problem is defined as searching over the sequence space to yield the maximum merit factor:
%
%
\quad $\frac{n^2}{2E(S)} \text{ with } E(S) = \sum_{k=1}^{n-1}\left(\sum_{i=1}^{n-k}s_i\cdot s_{i+k}\right)^2$. 
This hard, non-linear problem has been studied over several decades (see, e.g., \cite{LAC_Militzer,LABS_Packebusch2016}), where the only way to obtain exact solutions remains exhaustive search.
As a pseudo-Boolean function over $\{0,1\}^n$, it can be rewritten as follows:
%
%
\begin{equation}
    \displaystyle F_{\textrm{LABS}}\left(\vec{x}\right)= \frac{n^2} 
    {2 \sum\limits_{k=1}^{n-1}
    \left(\sum\limits_{i=1}^{n-k}
        x'_i \cdot x'_{i+k}
    \right)^{2}} \quad
    \textrm{where} \; x'_i = 2x_i - 1.
\end{equation}
\subsection{F19-F21: The Ising Model}
\label{sec:ising}
The Ising Spin Glass model \cite{Ising_Barahona1982} arose in solid-state physics and statistical mechanics, aiming to describe simple interactions within many-particle systems. 
The classical Ising model considers a set of spins placed on a regular lattice, where each edge $\left<i,j\right>$ is associated with an interaction strength $J_{i,j}$. In essence, a problem-instance is defined upon setting up the coupling matrix $\left\{J_{i,j} \right\}$. 
Each spin directs \textit{up} or \textit{down}, associated with a value $\pm 1$, and a set of $n$ spin glasses is said to form a configuration, denoted as $S=\left(s_1,\ldots,s_n\right)\in \{-1, +1\}^n$.
The configuration's energy function is described by the system's Hamiltonian, as a quadratic function of those $n$ spin variables: 
$ -\sum\limits_{i<j} J_{i,j}s_{i}s_{j} - \sum_{i=1}^{n} h_{i}s_{i} $, where $h_{i}$ is an external magnetic field. 
The optimization problem of interest is the study of the minimal energy configurations, which are termed \textit{ground states}, on a final lattice. This is clearly a challenging combinatorial optimization problem, which is known to be NP-hard, and to hold connections with all other NP problems \cite{Ising_Lucas2014}.
The Ising model has been investigated in light of EAs' operation, yielding some theoretical results on certain graph instances (see, e.g., \cite{Briest04,FischerWegener2005Ising,Sudholt2005Crossover}).

We have selected and integrated three Ising model instances in \tool, assuming zero external magnetic fields, and applying \emph{periodic boundary conditions} (PBC).
In order to formally define the Ising objective functions, we adopt a strict graph perspective, where $G=\left(V,E\right)$ is undirected and $V=[n]$. We apply an affine transformation $\left\{-1,+1\right\}^n \leadsto \left\{0,1\right\}^n$, where the $n$ spins become binary decision variables (this could be interpreted, e.g., as a coloring problem; see \cite{Sudholt2005Crossover}).
A generalized, compact form for the quadratic objective function is now obtained:
\begin{equation}\label{app:eq:compactIsing}
    F_{\textrm{Ising}}\left(\vec{x}\right)= \sum\limits_{\{u,v\}\in E} \left[x_{u}x_{v} - \left(1-x_{u} \right)\left(1-x_{v} \right) \right],
\end{equation}
thus leaving the instance definition within $G$.

In what follows, we specify their underlying graphs, whose edges are equally weighted as unity, to obtain their objective functions using \eqref{app:eq:compactIsing}.

\subsubsection{F19: The Ring (1D)}\label{sec:ising1}
This basic Ising model is defined over a one-dimensional lattice. The objective function follows \eqref{app:eq:compactIsing} using the following graph:
\begin{eqnarray}\nonumber
G_{\textrm{Is1D}}: & & \\
    e_{ij} = 1 & \Leftrightarrow & j = i+1 \; ~~~\forall i \in \{1,\ldots,n-1\}    \\ \nonumber
        & \vee & j=n,i=1
\end{eqnarray}
\subsubsection{F20: The Torus (2D)}\label{sec:ising2}
This instance is defined on a two-dimensional lattice of size $N$, using altogether $n=N^2$ vertices, denoted as $(i,j)$, $0\leq i,j \leq N-1$ \cite{Briest04}.
Since PBC are applied, a \emph{regular graph with each vertex having exactly four neighbors} is obtained.
The objective function follows \eqref{app:eq:compactIsing} using the following graph:
\begin{eqnarray}\nonumber
    G_{\textrm{Is2D}}: & & \\ \nonumber
    e_{(i,j)(k,\ell)} = 1 & \Leftrightarrow & \left[ k = (i+1)~\textrm{mod}~N ~~\wedge ~~ \ell=j ~~~\forall i,j \in \{0,\ldots,N-1\}  \right] \\ \nonumber
        & \vee &  \left[ k = (i-1)~\textrm{mod}~N ~~\wedge~~ \ell=j ~~~\forall i,j \in \{0,\ldots,N-1\}  \right] \\ \nonumber
        & \vee &  \left[\ell = (j+1)~\textrm{mod}~N ~~\wedge ~~k=i ~~~\forall j,i \in \{0,\ldots,N-1\}  \right]  \\ 
        & \vee &  \left[\ell = (j-1)~\textrm{mod}~N ~~\wedge ~~ k=i ~~~\forall j,i \in \{0,\ldots,N-1\}  \right]    \nonumber
\end{eqnarray}

\subsubsection{F21: Triangular (Isometric 2D Grid)}\label{sec:ising3}
This instance is also defined on a two-dimensional lattice, yet constructed on an isometric grid (also known as triangular grid), \emph{whose unit vectors form an angle of $\frac{2\pi}{3}$} \cite{Mellor_thesis}.
The vertices are placed on integer-valued two-dimensional $n=N^2$ vertices, denoted as $(i,j)$, $0\leq i,j \leq N-1$, yielding altogether a regular graph whose vertices have exactly six neighbors each (due to PBC):
\begin{eqnarray}\nonumber
    G_{\textrm{IsTR}}: & & \\ \nonumber
    e_{(i,j)(k,\ell)} = 1 & \Leftrightarrow & \left[ k = (i+1)~\textrm{mod}~N ~~\wedge ~~ \ell=j ~~~\forall i,j \in \{0,\ldots,N-1\}  \right] \\ \nonumber
        & \vee &  \left[ k = (i-1)~\textrm{mod}~N ~~\wedge~~ \ell=j ~~~\forall i,j \in \{0,\ldots,N-1\}  \right] \\ \nonumber
        & \vee &  \left[\ell = (j+1)~\textrm{mod}~N ~~\wedge ~~k=i ~~~\forall j,i \in \{0,\ldots,N-1\}  \right]  \\ \nonumber
        & \vee &  \left[\ell = (j-1)~\textrm{mod}~N ~~\wedge ~~ k=i ~~~\forall j,i \in \{0,\ldots,N-1\}  \right]  \\  \nonumber
        & \vee &  \left[\ell = (j+1)~\textrm{mod}~N ~~\wedge ~~k = (i+1)~\textrm{mod}~N ~~~\forall j,i \in \{0,\ldots,N-1\}  \right]  \\ \nonumber
        & \vee &  \left[\ell = (j-1)~\textrm{mod}~N ~~\wedge ~~ k = (i-1)~\textrm{mod}~N ~~~\forall j,i \in \{0,\ldots,N-1\}  \right]    \nonumber
\end{eqnarray}

\subsection{F22: Maximum Independent Vertex Set}\label{sec:MIVS}
Given a graph $G=([n],E)$, an independent vertex set is a subset of vertices where no two vertices are direct neighbors. 
A maximum independent vertex set (MIVS) (which generally is not equivalent to a \textit{maximal} independent vertex set) is defined as an independent subset $V^{\prime} \subset [n]$ having the largest possible size.
Using the standard binary encoding $V'=\{i \in [n] \mid x_i=1\}$, MIVS can be formulated as the maximization of the function 
\begin{equation}
     \displaystyle F_{\textrm{MIVS}}\left(x\right)= \sum\limits_{i} x_{i} - n\cdot \sum\limits_{i,j} x_{i}x_{j}e_{i,j},
 \end{equation}
where $e_{i,j}=1$ if $\{i,j\} \in E$ and $e_{i,j}=0$ otherwise.

In particular, following \cite{BK94}, we consider a specific, scalable problem instance, defining its Boolean graph as follows:
\begin{eqnarray}\nonumber
    e_{ij} = 1 & \Leftrightarrow & j = i+1 \; ~~~\forall i \in \{1,\ldots,n-1\}-\{n/2\}    \\
    & \vee & j=i+n/2+1 \; ~~~\forall i \in \{1,\ldots,n/2-1\} \\ \nonumber
    & \vee & j=i+n/2-1 \; ~~~\forall i \in \{2,\ldots,n/2\}.
\end{eqnarray}
The resulting graph has a simple, standard structure as shown in Figure \ref{app:fig:MIS-example} for $n=10$. 
The global optimizer has an objective function value of  $|V^{\prime}| = n/2+1$ for this standard graph.
Notably, $n \geq 4$ and $n$ is required to be even; given an odd $n$, we identify the $n$-dimensional problem with the $n-1$-dimensional instance. 

\begin{figure}[!ht]
  \centering
	\includegraphics[width=7cm]{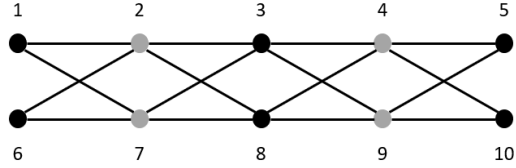}
  \caption{A scalable maximum independent set problem, with $n=10$ vertices and the optimal solution of size 6 marked by the black vertices.}
  \label{app:fig:MIS-example}
\end{figure}

\begin{figure}[t]
\centering
\includegraphics[width=0.7\linewidth]{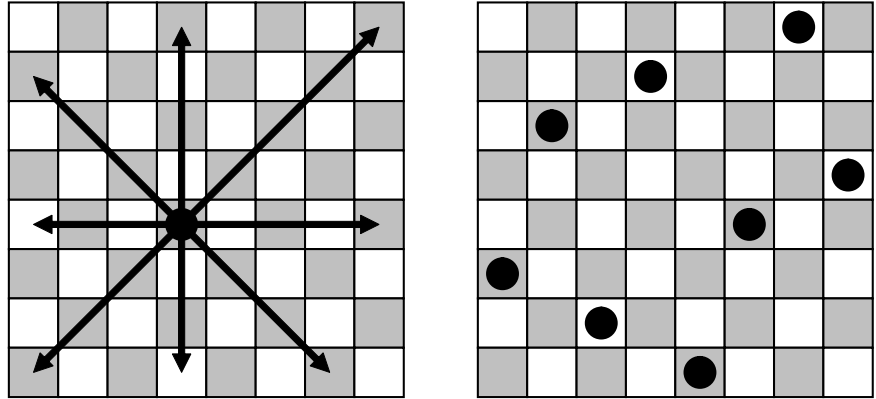}
\caption{The 8-queens problem: [Left] all possible fields a queen can move to from position D4; [Right] a feasible solution. \label{app:fig:8QP}} 
\end{figure}

\subsection{F23: N-Queens Problem}
\label{sec:Nqueens}
The $N$-queens problem (NQP) \cite{NQP_Bell2009} is defined as the task to place $N$ queens on an $N\times N$ chessboard in such a way that they cannot \textit{attack} each other.\footnote{The NQP is traced back to the 1848 Bezzel article entitled ``Proposal of the Eight Queens Problem''; for a comprehensive list of references we refer the reader to a documentation by W.~Kosters at\\ \url{http://liacs.leidenuniv.nl/~kosterswa/nqueens/nqueens_feb2009.pdf}. }
Figure \ref{app:fig:8QP} provides an illustration for the 8-queens problem. 
Notably, the \emph{NQP is actually an instance of the MIVS problem} -- when considering a graph on which all possible queen-attacks are defined as edges.
NQP formally constitutes a \textit{Constraints Satisfaction Problem}, but is posed here as a maximization problem using a binary representation:
$$\begin{array}{l}
\textrm{maximize} ~ \sum\limits_{i,j} x_{ij}\\
\textrm{subject to:}\\
\displaystyle~~~~~\sum\limits_{i} x_{ij} \leq 1 ~~~ \forall j \in \left\{1\ldots,N \right\} \\
\displaystyle~~~~~\sum\limits_{j} x_{ij} \leq 1 ~~~ \forall i \in \left\{1\ldots,N \right\} \\
\displaystyle~~~~~\sum\limits_{j-i=k} x_{ij} \leq 1 ~~~ \forall k \in \left\{-N+2,-N+3,\ldots,N-3,N-2 \right\} \\
\displaystyle~~~~~\sum\limits_{i+j=\ell} x_{ij} \leq 1 ~~~ \forall \ell \in \left\{3,4,\ldots,2N-3,2N-1 \right\} \\
\displaystyle\displaystyle~~~~~x_{ij}\in\left\{0,1\right\} ~ \forall i,j \in \left\{1,\ldots,N \right\}
\end{array}
$$
This formulation utilizes $n=N^2$ binary decision variables $x_{ij}$, which are associated with the chessboard's coordinates, having an origin $(1,1)$ at the top-left corner. Setting a binary to $1$ implies a single queen assignment in that cell. This formulation promotes placement of as many queens as possible by means of the objective function, followed by four sets of constraints eliminating queens' \textit{mutual threats}: the first two sets ensure a single queen on each row and each column, whereas the following two sets ensure a single queen at the increasing-diagonals (using the dummy indexing $k$) and decreasing-diagonals (using the dummy indexing $\ell$). 
It should be noted that a \textit{permutation formulation} also exists for this problem, and is sometimes attractive for RSHs.
Due to chessboard symmetries, NQP possesses multiplicity of optimal solutions. 
Its attractiveness, however, lies in its hardness. 
In terms of a black-box objective function, we formulate NQP as the maximization of the following function:
\begin{equation}
\begin{aligned}
    \displaystyle F_{\textrm{NQP}}(\vec{x}) & =  \sum_{i=1}^{N}\sum_{j=1}^{N} x_{ij} - N\cdot \left( \sum_{i=1}^{N}\max\left\{ 0, -1 +\sum_{j=1}^{N} x_{ij} \right\} +
    \sum_{j=1}^{N}\max\left\{ 0, -1 +\sum_{i=1}^{N} x_{ij} \right\}\right. \\
    & +   \! \! \! \!
    \left. \sum\limits_{k=-N+2}^{N-2}\max\left\{0, -1+ \! \! \! \! \! \! \! \! \sum\limits_{{j-i=k}  \atop {i,j \in \{1,2,\ldots,N \}}} \! \! \! \! \! \! \! \!  x_{ij}\right\} +
    \sum\limits_{\ell=3}^{2N-1} \max\left\{0, -1+ \! \! \! \! \! \! \! \! \sum\limits_{{j+i=\ell} \atop {i,j \in \{1,2,\ldots,N \}}} \! \! \! \! \! \! \! \!  x_{ij}\right\}\right) 
\end{aligned}
\end{equation}

\section{Algorithms}
\label{sec:algos}

We evaluate a total number of twelve different algorithms on the problems described above. We have chosen algorithms that may serve for future references, since they all have some known strengths and weaknesses that will become apparent in the following discussions. Our selection therefore shows a clear bias towards algorithms for which theoretical analyses are available. 

Note that most algorithms are parametrized, and we use here in this work only standard parametrizations (e.g., we use $1/n$ as mutation rates, etc.). Analyzing the effects of different parameter values as was done, for example in~\cite{Arina2019,DangD19}, would be very interesting, but is beyond the scope of this present work. 

We also note that, except for the so-called vGA, our implementations (deliberately) deviate slightly from the text-book descriptions referenced below. Following the suggestions made in~\cite{CarvalhoD17}, we enforce that offspring created by mutation are different from their parent and resample without further evaluation if needed. Likewise, we do not evaluate recombination offspring that are identical to one of their immediate parents. Since we use this convention throughout, we omit the subscript $_{>0}$ used in~\cite{CarvalhoD17,DoerrYRWB18}. 

All algorithms start with uniformly chosen initial solution candidates. 

We list here the twelve implemented algorithms, and provide further details and pseudo-codes thereafter:
\begin{enumerate}
    \item \textbf{gHC:} A (1+1) greedy hill climber, which goes through the string from left to right, flipping exactly one bit per each iteration, and accepting the offspring if it is at least as good as its parent. 
    \item \textbf{RLS:} Randomized Local Search, the elitist (1+1) strategy flipping one uniformly chosen bit in each iteration. That is, RLS and gHC differ only in the choice of the bit which is flipped. While RLS is unbiased in the sense of Section~\ref{sec:unbiasedness}, gHC is not permutation-invariant and thus biased.
    \item \textbf{$(1+1)$ EA:} The (1+1) EA with static mutation rate $p=1/n$. This algorithm differs from RLS in that the number of uniformly chosen, pairwise different bits to be flipped is sampled from the conditional binomial distribution $\Bin_{>0}(n,p)$. That is, each bit is flipped independently with probability $p$ and if none of the bits has been flipped this \emph{standard bit mutation} is repeated until we obtain an offspring that differs from its parent in at least one bit. 
    \item \textbf{fGA:} The ``fast GA'' proposed in~\cite{FastGA17} with $\beta=1.5$. Its \emph{mutation strength} (i.e., the number of bits flipped in each iteration) follows a power-law distribution with exponent $\beta$. This results in a more frequent use of large mutation-strength, while maintaining the property that small mutation strengths are still sampled with reasonably large probability.
    \item \textbf{$(1+10)$ EA:} The (1+10)~EA with static $p=1/n$, which differs from the (1+1) EA only in that 10 offspring are sampled (independently) per each iteration. Only the best one of these (ties broken at random) replaces the parent, and only if it is as least as good.  
    \item \textbf{$(1+10)$ EA$_{r/2,2r}$:} The two-rate EA with self-adjusting mutation rates suggested and analyzed in~\cite{DoerrGWY17}. 
    \item \textbf{$(1+10)$ EA$_{\norm}$:} a variant of the $(1+10)$~EA sampling the mutation strength from a normal distribution $N(pn,pn(1-p))$ with a self-adjusting choice of $p$~\cite{YeDB19}.
    \item \textbf{$(1+10)$ EA$_{\vars}$:} The $(1+10)$ EA$_{\norm}$ with an adaptive choice of the variance in the normal distribution from which the mutation strengths are sampled. Also from~\cite{YeDB19}.
    \item \textbf{$(1+10)$ EA$_{\lognormal}$} The (1+10) EA with log-normal self-adaptation of the mutation rate proposed in~\cite{BackS96}.
    \item \textbf{$(1+(\lambda,\lambda))$~GA:} A binary (i.e., crossover-based) EA originally suggested in~\cite{DoerrDE15}. We use the variant with self-adjusting $\lambda$ analyzed in~\cite{DoerrD18ga}.
    \item \textbf{vGA:} A $(30,30)$ ``vanilla'' GA (following the so-called traditional GA, as described, for example, in~\cite{Goldberg,Baeck-book}).
    \item \textbf{UMDA:} A univariate marginal distribution algorithm from the family of estimation of distribution algorithms (EDAs). UMDA was originally proposed in~\cite{Muhlenbein97}.
\end{enumerate}

\subsection{Detailed Description of the Algorithms}
A detailed description of the algorithms follows. An operator frequently used in these descriptions is the $\mut_{\ell}(\cdot)$ mutation operator, which flips the entries of $\ell$ pairwise different, uniformly at random chosen bit positions, see Algorithm~\ref{alg:mut}. 

\begin{algorithm2e}%
	\textbf{Input:} $x \in \{0,1\}^n$, $\ell \in \N$\;
		\label{line:elloea}Select $\ell$ pairwise different positions $i_1,\ldots,i_{\ell} \in [n]$ uniformly at random\;
	  $y \assign x$\;
		\lFor{$j=1,...,\ell$}{$y_{i_j}\assign 1-x_{i_j}$}
\caption{$\mut_{\ell}$ chooses $\ell$ different positions and flips the entries in these positions.}
\label{alg:mut}
\end{algorithm2e}

It is well known that standard bit mutation, which flips each bit in a length-$n$ bit string with some probability $p$, can be equivalently defined as the operator calling $\mut_{\ell}$ for a \emph{mutation strength} chosen from the binomial distribution $\Bin(n,p)$. Since we want to avoid useless evaluations of offspring  that are identical to their parents, we frequently make use of the conditional binomial distribution $\Bin_{>0}(n,p)$, which assigns probability $\Bin(n,p)(k)/(1-(1-p)^n)$ to each positive integer $k \in [n]$, and probability zero to all other values. Sampling from $\Bin_{>0}(n,p)$ is identical to sampling from $\Bin(n,p)$ until a positive value is returned (``resampling strategy'').
%


\subsubsection{Greedy Hill Climber}
\label{sec:gHC}

The greedy hill climber (gHC, Algorithm~\ref{alg:gHC}) uses a deterministic mutation strength, and flips one bit in each iteration, going through the bit string from left to right, until being stuck in a local optimum, see Algorithm~\ref{alg:gHC}. 

\begin{algorithm2e}[H]
\textbf{Initialization:} 
	Sample $x \in \{0,1\}^{n}$ uniformly at random and evaluate $f(x)$\;
	\textbf{Optimization:}
	\For{$t=1,2,3,\ldots$}{
	    $x^* \assign x$\; 
	    Flip in $x^*$ the entry in position $1+(t \mod n)$ 
	    and evaluate $f(x^*)$\;
		\lIf{$f(x^*)\ge f(x)$}{$x \assign x^*$}
	}
\caption{Greedy hill climber (gHC)}
\label{alg:gHC}
\end{algorithm2e}


\subsubsection{Randomized Local Search}
\label{sec:RLS}

RLS uses a deterministic mutation strength, and flips one randomly chosen bit in each iteration, see Algorithm~\ref{alg:RLS}. 

\begin{algorithm2e}[H]
\textbf{Initialization:} 
	Sample $x \in \{0,1\}^{n}$ uniformly at random and evaluate $f(x)$\;
	\textbf{Optimization:}
	\For{$t=1,2,3,\ldots$}{
	    create $x^* \assign \mut_{1}(x)$, and evaluate $f(x^*)$\;
		\lIf{$f(x^*)\ge f(x)$}{$x \assign x^*$}
	}
\caption{Randomized local search (RLS)}
\label{alg:RLS}
\end{algorithm2e}

\subsubsection{The EA with Static Mutation Rate}
\label{sec:opl}

The $(1+\lambda)$ EA is defined via the pseudo-code in Algorithm~\ref{alg:opl}. We use $\lambda=1$ and $\lambda=10$ in our comparisons.

\begin{algorithm2e}[H]
\textbf{Initialization:} 
	Sample $x \in \{0,1\}^{n}$ uniformly at random and evaluate $f(x)$\;
	\textbf{Optimization:}
	\For{$t=1,2,3,\ldots$}{
	    \For{$i=1,\ldots,\lambda$}{
	        Sample $\ell^{(i)} \sim \Bin_{>0}(n,1/n)$\;
	        create $y^{(i)} \assign \mut_{\ell^{(i)}}(x)$, and evaluate $f(y^{(i)})$\;
	        }
	        $x^* \assign \arg\max\{f(y^{(1)}), \ldots, f(y^{(\lambda)})\}$ (ties broken by selecting the first max $f(y^{(i)})$)\;
		    \lIf{$f(x^*)\ge f(x)$}{$x \assign x^*$}

	}
\caption{The $(1+\lambda)$~EA with static mutation rates}
\label{alg:opl}
\end{algorithm2e}

\subsubsection{Fast Genetic Algorithm}
\label{sec:fastGA}

The \emph{fast Genetic Algorithm} (fGA) chooses the mutation length $\ell$ according to a power-law distribution $D_{n/2}^{\beta}$, which assigns to each  integer $k \in [n/2]$ a probability of $\Pr[D_{n/2}^{\beta}=k]={(C_{n/2}^{\beta})}^{-1}k^{-\beta}$, where $C_{n/2}^{\beta} = \sum_{i=1}^{n/2}i^{-\beta}$. We use the (1+1) variant of this algorithm with $\beta=1.5$.

\begin{algorithm2e}[H]
\textbf{Initialization:} 
	Sample $x \in \{0,1\}^{n}$ uniformly at random and evaluate $f(x)$\;
	\textbf{Optimization:}
	\For{$t=1,2,3,\ldots$}{
	    \For{$i=1,\ldots,\lambda$}{
	        Sample $\ell^{(i)} \sim D_{n/2}^{\beta}$\;
	        create $y^{(i)} \assign \mut_{\ell^{(i)}}(x)$, and evaluate $f(y^{(i)})$\;
	        }
	        $x^* \assign \arg\max\{f(y^{(1)}), \ldots, f(y^{(\lambda)})\}$ (ties broken by favoring the largest index)\;
		    \lIf{$f(x^*)\ge f(x)$}{$x \assign x^*$}

	}
\caption{Fast genetic algorithm (fGA) from \cite{FastGA17}}
\label{alg:fGA}
\end{algorithm2e}


\subsubsection{The Two-Rate EA}
The two-rate $(1+\lambda)$~EA$_{r/2,2r}$ was introduced in \cite{DoerrGWY17}. It uses two mutation rates in each iteration: half of offspring are generated with mutation rate $r/2n$, and the other $\lambda/2$ offspring are sampled  with mutation rate $2r/n$. The parameter $r$ is updated after each iteration, from a biased random decision favoring the value from which the best of all $\lambda$ offspring has been sampled.

\begin{algorithm2e}[H]
	\textbf{Initialization:} 
	Sample $x \in \{0,1\}^{n}$ uniformly at random and evaluate $f(x)$\;
	Initialize $r \assign r^{\text{init}}$; // we use $r^{\text{init}}=2$\;
  \textbf{Optimization:}
	\For{$t=1,2,3,\ldots$}{
		\For{$i=1,\ldots,\lambda/2$}{
			\label{line:half}Sample $\ell^{(i)} \sim \Bin_{>0}(n,r/(2n))$, 
			create $y^{(i)} \assign \mut_{\ell^{(i)}}(x)$, and
			evaluate $f(y^{(i)})$\;
		 }
		\For{$i=\lambda/2+1,\ldots,\lambda$}{
			\label{line:double}Sample $\ell^{(i)} \sim \Bin_{>0}(n,2r/n)$, 
			create $y^{(i)} \assign \mut_{\ell^{(i)}}(x)$, and
			evaluate $f(y^{(i)})$\;
		 }
		$x^* \assign \arg\max\{f(y^{(1)}), \ldots, f(y^{(\lambda)})\}$ (ties broken uniformly at random)\;
		\lIf{$f(x^*)\ge f(x)$}{$x \assign x^*$}
		\lIf{$x^{*}$ \text{has been created with mutation rate} $r/2$}{$s \assign 3/4$ \textbf{ else} $s \assign 1/4$}
		Sample $q \in [0,1]$ uniformly at random\;
		\lIf{$q\le s$}{$r \assign \max\{r/2,2\}$ \textbf{ else} {$r \assign \min\{2r,n/4\}$}} 
	}
\caption{The two-rate EA with adaptive mutation rates proposed in~\cite{DoerrGWY17}}
\label{alg:2rateEA}
\end{algorithm2e}


\subsubsection{The EA with normalized standard bit mutation}
\label{sec:EAnorm}

The $(1+\lambda)$~EA$_{\text{norm.}}$, Algorithm~\ref{alg:EAnorm}, samples the mutation strength from a  normal distribution with mean $r=pn$ and variance $pn(1-p)=r(1-r/n)$, which is identical to the variance of the binomial distribution used in standard bit mutation. The parameter $r$ is updated after each iteration, in a similar fashion as in the 2-rate EA, Algorithm~\ref{alg:2rateEA}.

 \begin{algorithm2e}[H]%
	\textbf{Initialization:} 
	Sample $x \in \{0,1\}^{n}$ uniformly at random and evaluate $f(x)$\;
	Initialize $r \assign r^{\text{init}}$; // we use $r^{\text{init}}=2$\;
  \textbf{Optimization:}
	\For{$t=1,2,3,\ldots$}{
		\For{$i=1,\ldots,\lambda$}{
			\label{line:mutnormal}Sample 
			$\ell^{(i)} \sim \min\{N_{>0}(r,r(1-r/n)),n\}$, 
			create $y^{(i)} \assign \mut_{\ell^{(i)}}(x)$, and
			evaluate $f(y^{(i)})$\;
		 }
		$i \assign \min\left\{ j \mid f(y^{(j)}) = \max\{f(y^{(k)}) \mid k \in [\lambda]\} \right\}$\;
				$r \assign \ell^{(i)}$\; 
		\lIf{$f(y^{(i)})\ge f(x)$}{$x \assign y^{(i)}$}
	}
\caption{The $(1+\lambda)$~EA$_{\text{norm.}}$ with normalized standard bit mutation}
\label{alg:EAnorm}
\end{algorithm2e}

\subsubsection{The EA with normalized standard bit mutation and controlled variance}
\label{sec:EAvars}

The $(1+\lambda)$~EA$_{\text{var.}}$, Algorithm~\ref{alg:EAvars} builds on the $(1+\lambda)$~EA$_{\text{norm.}}$ but uses, in addition to the adaptive choice of the mutation rate, an adaptive choice of the variance.

 \begin{algorithm2e}[H]%
	\textbf{Initialization:} 
	Sample $x \in \{0,1\}^{n}$ uniformly at random and evaluate $f(x)$\;
	Initialize $r \assign r^{\text{init}}$; // we use $r^{\text{init}}=2$\;
	Initialize $c \assign 0$\;
  \textbf{Optimization:}
	\For{$t=1,2,3,\ldots$}{
		\For{$i=1,\ldots,\lambda$}{
			\label{line:mutnormaladap}Sample 
			$\ell^{(i)} \sim \min\{N_{>0}(r,F^c r(1-r/n)),n\}$, 
			create $y^{(i)} \assign \mut_{\ell^{(i)}}(x)$, and
			evaluate $f(y^{(i)})$\;
		 }
		$i \assign \min\left\{ j \mid f(y^{(j)}) = \max\{f(y^{(k)}) \mid k \in [\lambda]\} \right\}$\;
		\lIf{$r =\ell^{(i)}$}{$c \assign c+1$; \textbf{else} $c \assign 0$}
		$r \assign \ell^{(i)}$\; 
		\lIf{$f(y^{(i)})\ge f(x)$}{$x \assign y^{(i)}$}
	}
\caption{The $(1+\lambda)$~EA$_{\text{var.}}$ with normalized standard bit mutation and a self-adjusting choice of mean and variance}
\label{alg:EAvars}
\end{algorithm2e}

\subsubsection{The EA with log-Normal self-adaptation on mutation rate}
\label{sec:EAlognormal}

The $(1+\lambda)$~EA$_{\lognormal}$, Algorithm~\ref{alg:EAlognormal}, uses a self-adaptive choice of the mutation rate.

\begin{algorithm2e}[H]
\textbf{Initialization:} 
	Sample $x \in \{0,1\}^{n}$ uniformly at random and evaluate $f(x)$\;
	$p=0.2$\;
	\textbf{Optimization:}
	\For{$t=1,2,3,\ldots$}{
	    \For{$i=1,\ldots,\lambda$}{
	        $p^{(i)} = \big(1+\frac{1-p}{p}\cdot \exp(0.22\cdot \mathcal{N}(0,1))\big)^{-1}$ \;
	        Sample $\ell^{(i)} \sim \Bin_{>0}(n,p^{(i)})$\;
	        create $y^{(i)} \assign \mut_{\ell^{(i)}}(x)$, and evaluate $f(y^{(i)})$\;
	        }
	    $i \assign \min\left\{ j \mid f(y^{(j)}) = \max\{f(y^{(k)}) \mid k \in [\lambda]\} \right\}$\;
		$p \assign p^{(i)}$\; 
	    $x^* \assign \arg\max\{f(y^{(1)}), \ldots, f(y^{(\lambda)})\}$ (ties broken by favoring the smallest index)\;
		\lIf{$f(x^*)\ge f(x)$}{$x \assign x^*$}
	}
\caption{The $(1+\lambda)$~EA$_{\lognormal}$ with log-Normal self-adaptation of the mutation rate}
\label{alg:EAlognormal}
\end{algorithm2e}

\subsubsection{The Self-Adjusting \ga}
\label{sec:ga}

The self-adjusting $(1+(\lambda,\lambda))$ GA, Algorithm~\ref{alg:ga}, was introduced in~\cite{DoerrDE15} and analyzed in~\cite{DoerrD18ga}. The offspring population size $\lambda$ is updated after each iteration, depending on whether or not an improving offspring could be generated. Since both the mutation rate and the crossover bias (see Algorithm~\ref{alg:cross} for the definition of the biased crossover operator $\cross$) depend on $\lambda$, these two parameters also change during the run of the \ga. In our implementation we use update strength $F=3/2$.

\begin{algorithm2e}[H]
\textbf{Initialization:} 
	Sample $x \in \{0,1\}^{n}$ uniformly at random and evaluate $f(x)$\;
\textbf{Optimization:}
	\For{$t=1,2,3,\ldots$}{
	    \textbf{Mutation phase:}\\
			\Indp
	    Sample $\ell \sim \Bin_{>0}(n,\lambda/n)$\;  
	    \lFor{$i=1,\ldots,\lambda$}{
	        create $y^{(i)} \assign \mut_{\ell}(x)$, and evaluate $f(y^{(i)})$
	    }
	    $x^* \assign \arg\max\{f(y^{(1)}), \ldots, f(y^{(\lambda)})\}$ (ties broken by favoring the largest index)\;
			\Indm
		\textbf{Crossover phase:}\\
		\Indp
		\lFor{$i=1,\ldots,\lambda$}{
	        create $y^{(i)} \assign \cross_c(x,x^*)$, and evaluate $f(y^{(i)})$
	    }
		$y^* \assign \arg\max\{f(y^{(1)}), \ldots, f(y^{(\lambda)})\}$ (ties broken by favoring the largest index)\;
		\Indm
		\textbf{Selection phase:}\\
		\Indp
        \lIf{$f(y^*) > f(x)$}{$x \assign y^*$; $\lambda \assign \max\{\lambda/F,1\}$}
        \lIf{$f(y^*) = f(x)$}{$x \assign y^*$; $\lambda \assign \min\{\lambda F^{1/4},n\}$}
         \lIf{$f(y^*) < f(x)$}{$\lambda \assign\min\{\lambda F^{1/4},n\}$}
		\Indm		
	}
\caption{The self-adjusting \ga}
\label{alg:ga}
\end{algorithm2e}

\begin{algorithm2e}
$y \assign x$\;
Sample $\ell \sim \Bin_{>0}(n,c)$\;
Select $\ell$ different positions $\{i_1, \ldots, i_l \} \in [n]$\;
\lFor{$j=1,2,\ldots,\ell$}{
    $y_{i_j} \assign x^*_{i_j}$}
\caption{Crossover operation $\cross_c(x,x^*)$ with crossover bias $c$}
\label{alg:cross}
\end{algorithm2e}


\subsubsection{The ``Vanilla'' GA}
\label{sec:vGA}

The vanilla GA (vGA, Algorithm~\ref{alg:vGA}) constitutes a textbook realization of the so-called Traditional GA~\cite{Goldberg,Baeck-book}. 
The algorithm holds a parental population of size $\mu$. It employs the Roulette-Wheel-Selection (RWS, that is, probabilistic fitness-proportionate selection which permits an individual to appear multiple times) as the sexual selection operator to form $\mu/2$ pairs of individuals that generate the offspring population.  
1-point crossover (Algorithm~\ref{alg:1pt}) is applied to every pair with a fixed probability of $p_c=0.37$. A mutation operator is then applied to every individual, flipping every bit with a fixed probability of $p_m = 2/n$.
This completes a single cycle.

\begin{algorithm2e}[t]
Sample $\ell \in [n]$ uniformly at random\;
\lFor{$i=1,2,\ldots,\ell$}{
    Set $y^{(1)}_{i} \assign x^{(1)}_i$ and $y^{(2)}_{i} \assign x^{(2)}_i$}
\lFor{$i=\ell+1,\ldots,n$}{
    Set $y^{(1)}_{i} \assign x^{(2)}_i$ and $y^{(2)}_{i} \assign x^{(1)}_i$}
\caption{1-Point crossover of two parents $x^{(1)}$ and $x^{(2)}$}
\label{alg:1pt}
\end{algorithm2e}

\begin{algorithm2e}[H]%
    \textbf{Initialization:}\\
    \Indp
    \lFor{$i=1,\ldots,\mu$}{sample $x^{(i)} \in \{0,1\}^n$ uniformly at random and evaluate $f(x^{(i)})$}
    \Indm
	\textbf{Optimization:} ~ \For{$t=1,2,3,\ldots$}{
	    \textbf{Parent selection phase:} Apply roulette-wheel selection to $\{x^{(1)}, \ldots, x^{\mu}\}$ to select $\mu$ parent individuals $y^{(1)}, \ldots, y^{(\mu)}$\;
		\Indp
		\Indm
		\textbf{Crossover phase:}\\
		\Indp
		    \lFor{$i=1,\ldots,\mu/2$}{
	        with probability $p_c$ replace $y^{(i)}$ and $y^{(2i)}$ by the two offspring that result from a 1-point crossover of these two parents, for a randomly chosen crossover point $j \in [n]$ 
		    }
		\Indm
		\textbf{Mutation phase:}\\
			\Indp
		    \lFor{$i=1,\ldots,\mu$}{
	        Sample $\ell^{(i)} \sim \Bin(n,p_m)$, set 
	        $y^{(i)} \assign \mut_{\ell^{(i)}}(y^{(i)})$, and evaluate $f(y^{(i)})$
	        }
		    \Indm
		\textbf{Replacement:}\\
			\Indp
		    \lFor{$i=1,\ldots,\mu$}{Replace $x^{(i)}$ by $y^{(i)}$}
	}
\caption{The $(\mu,\mu)$-``Vanilla-GA'' with mutation rate $p_m$ and crossover probability $p_c$}
\label{alg:vGA}
\end{algorithm2e}


\subsubsection{The Univariate Marginal Distribution Algorithm}
\label{sec:UMDA}

The univariate marginal distribution algorithm (UMDA, Algorithm~\ref{alg:UMDA}) is one of the simplest representatives of the family of so-called estimation of distribution algorithms (EDAs). The algorithm maintains a population of size $s$ (we use $s=50$ in our experiments) and uses the best $s/2$ of these to estimate the marginal distribution of each decision variable, by simply counting the relative frequency of ones in the corresponding position. These frequencies are capped at $1/n$ and $1-1/n$, respectively. In the $t$-th iteration, a new population is created by sampling from these marginal distributions. Building upon previous work made in~\cite{MuhlenbeinV96}, the UMDA was introduced in~\cite{Muhlenbein97}. Theoretical results for this algorithm are summarized in~\cite{KrejcaW18}.

\begin{algorithm2e}%
	 \textbf{Initialization:}\\
    \Indp
    \lFor{$i=1,\ldots,s$}{sample $x^{(0,i)} \in \{0,1\}^n$ uniformly at random and evaluate $f(x^{(0,i)})$}
		Let $\vec{P_0}$ be the collection of the best $s/2$ of these search points, ties broken uniformly at random (u.a.r.)\; 
    \Indm
  \textbf{Optimization:}\\
	 \Indp
	\For{$t=1,2,3,\ldots$}{
		\For{$j=1,\ldots,n$}{
		$p_j \leftarrow 2|\{ x \in \vec{P_{t-1}} \mid x_j=1 \}|/s$\;
		\lIf{$p_j < 1/n$}{$p_j = 1/n$}
		\lIf{$p_j > 1 - 1/n$}{$p_j = 1 - 1/n$}
		}
		\lFor{$i=1,\ldots,s$}{
		sample $x^{(t,i)} \in \{0,1\}^n$ by setting, independently for all $j \in [n]$, $x^{(t,i)}_j=1$ with probability $p_j$ and setting $x^{(t,i)}_j=0$ otherwise. Evaluate $f(x^{(t,i)})$}
		Let $\vec{P_t}$ be the collection of the best $s/2$ of the points $x^{(t,1)},x^{(t,2)},..., x^{(t,s)}$, ties broken u.a.r.\;
	}
	  \Indm
\caption{The Univariate Marginal Distribution Algorithm (UMDA), representing the family of EDAs}
\label{alg:UMDA}
\end{algorithm2e}

\section{Experimental Results}
\label{sec:experiments}
As a demonstration of the results that can be obtained from the \tool environment, we report here on some basic experiments that were run on it. All data is available for interactive evaluation with \textsc{IOHanalyzer} at \url{http://iohprofiler.liacs.nl} (the data sets can be loaded by selecting the data set 2019gecco-inst11-1run in the ``Load Data from Repository'' section on the ``upload data'' page. The data set with 11 runs on the first instance is available as data set 2019gecco-inst1-11run). In addition to analyzing the performance statistics of the baseline algorithms described above, the user can upload his/her own data sets for a performance comparison against the twelve algorithms, or for an extension of this work to more benchmark problems.

Following the primary scope of this work, which is a demonstration of the ability of \tool to handle such large-scale experiments, our report features only the major empirical findings. A more detailed analysis by each algorithm is left for future work. We note, though, that the insights obtained through the here-described experiments have already inspired the development of new algorithmic ideas, see~\cite{YeDB19,Arina2019,HoreshBS19}.

\begin{figure*}[t]
    \centering
    \includegraphics[width=0.96\linewidth]{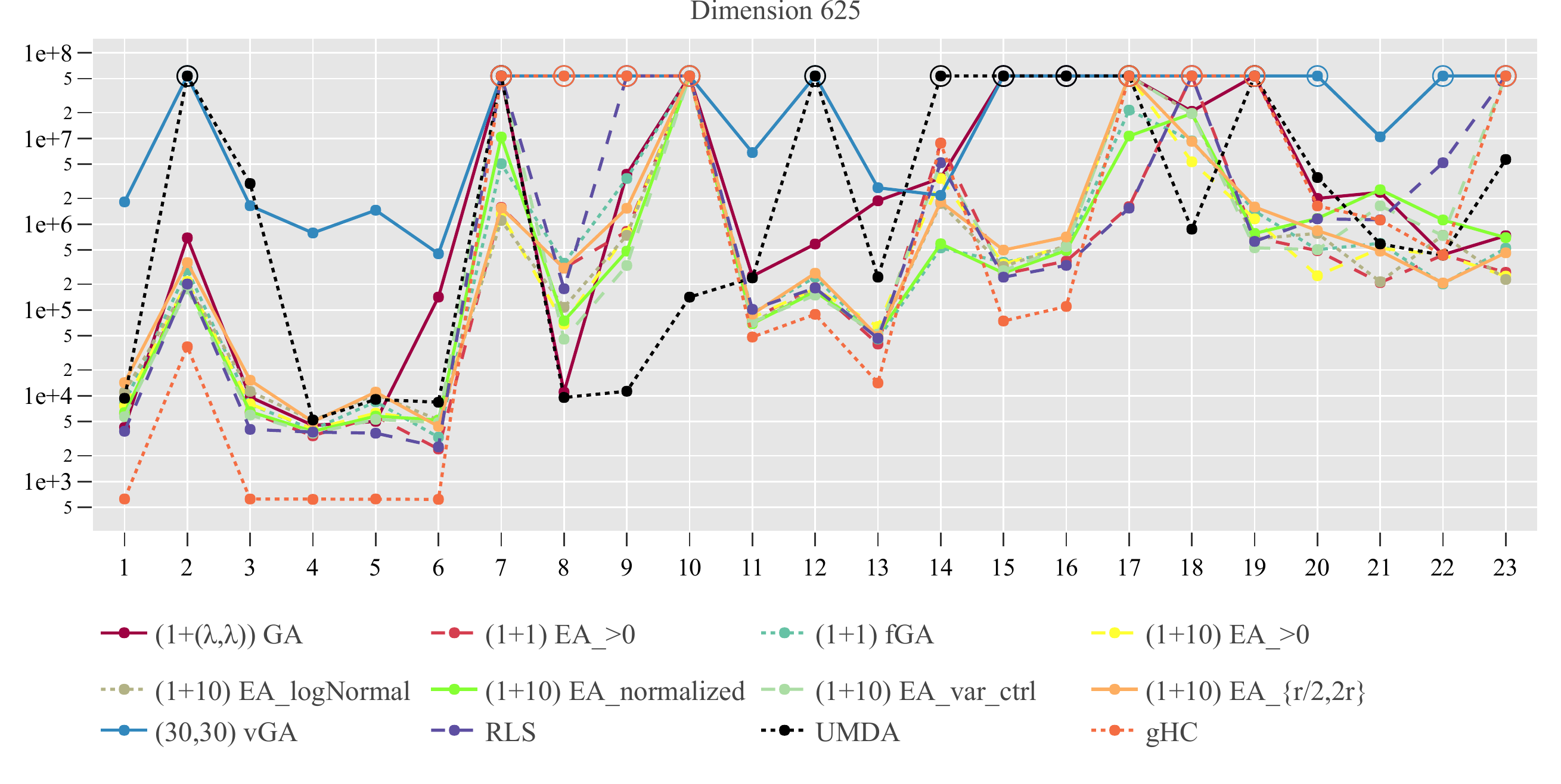}
    \caption{ERT values of the twelve baseline algorithms for the 625-dimensional test suite, with respect to the best solution quality found by any of the algorithms in any of the eleven runs. These target values can be found in Table~\ref{tab:targets}.
		\label{fig:ERT-625}}
\end{figure*}

\begin{figure*}
    \centering
    \includegraphics[width=0.96\linewidth]{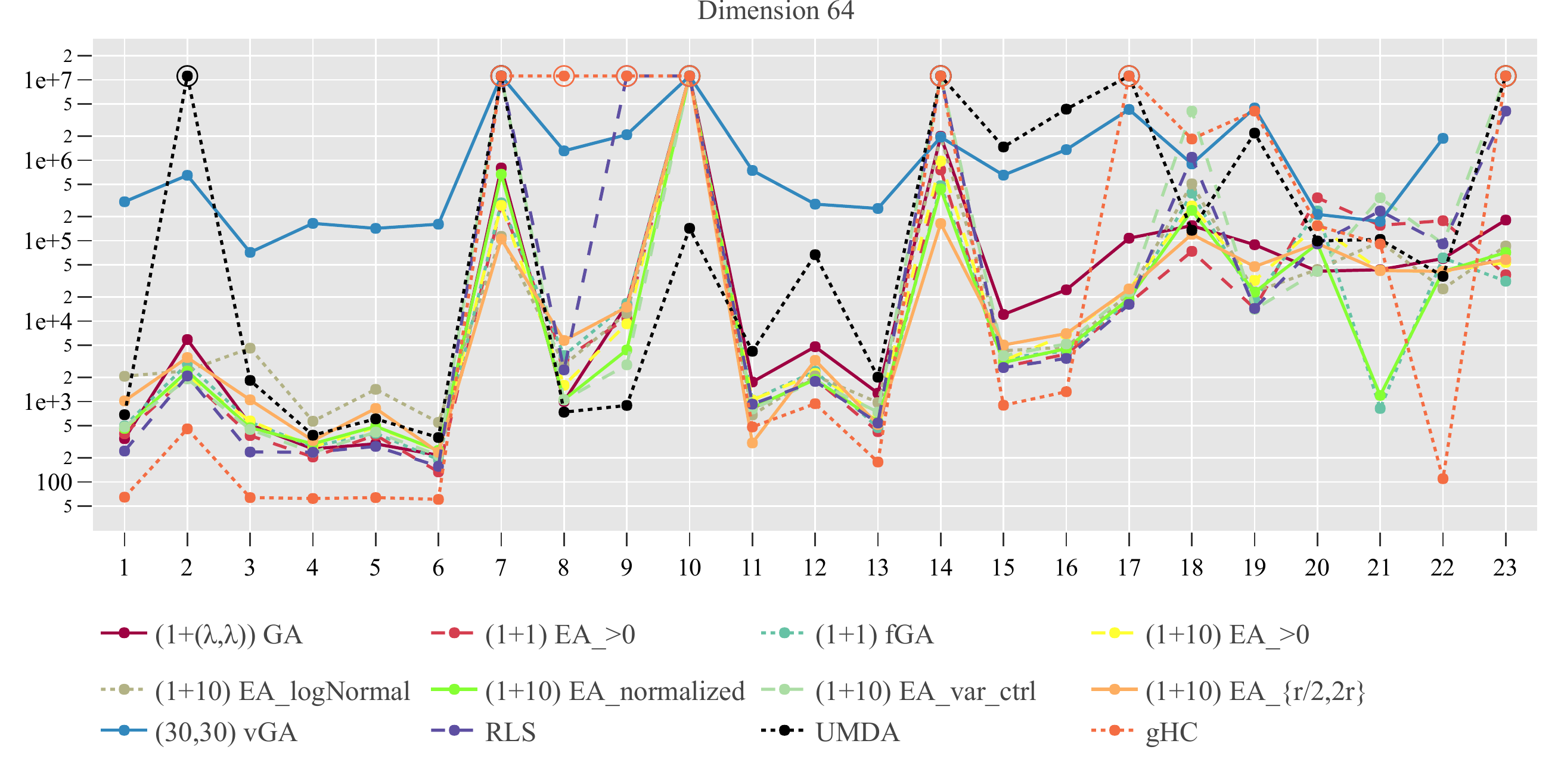}
    \caption{ERT values of the twelve baseline algorithms for the 64-dimensional test suite, with respect to the best solution quality found by any of the algorithms in any of the eleven runs. These target values can be found in Table~\ref{tab:targets}.
		\label{fig:ERT-64}}
\end{figure*}

\subsection{Experimental Setup}
\label{sec:setup}

Our experimental setup can be summarized as follows:
\begin{itemize}
\item 23 test-functions F1-F23, described in Section~\ref{sec:functions}
\item Each function is assessed over the four problem dimensions $n \in \left\{16, 64, 100, 625 \right\}$
\item Each algorithm is run on 11 different instances of each of these 92 $(F,n)$ pairs, yielding a total number of 1,012 different runs per each algorithm. Each run is granted a budget of $100n^2$ function evaluations for dimensions $n\in\{16,64,100\}$ and a budget of $5 n^2$ function evaluations for $n=625$. More precisely, each algorithm performs one run on each of the instances $1-6$ and $51-55$ described in Section~\ref{sec:unbiasedness}.

As mentioned in Section~\ref{sec:algos}, most of our algorithms are unbiased and comparison-based. For these algorithms all 11 instances look the same, i.e., performing one run each is equivalent to 11 independent runs on instance 1, which is the ``pure'' problem instance without fitness scaling nor any other transformation applied to it. However, in order to understand how the transformations impact the behavior of vGA and gHC, we also performed 11 independent runs of each algorithm on instance 1 of each $(F,n)$ pair, yielding another 1,012 runs per each algorithm.
\item For each run we store the current and the best-so-far function value at each evaluation. This setup allows very detailed analyses, since we can zoom into each range of fixed budgets and/or fixed-targets of choice, and obtain our anytime performance statistics in terms of quantiles, averages, probabilities of success, ECDF curves, etc. 

 For some of the algorithms we also store information about the self-adjusting parameters, for example the value of $\lambda$ in the \ga and the mutation rates for the (1+10) EA$_{r/2,2r}$, the (1+10) EA$_{\vars}$, and the (1+10) EA$_{\norm}$. From this data we can derive how the parameters evolve with respect to the time elapsed and with respect to the quality of the best-so-far solutions.
\end{itemize}

As discussed, we are interested in experimental setups that allow to evaluate one entire experiment (for one algorithm) within \textbf{24 CPU hours.} The majority of our tested algorithms completed the experiment in around 12 CPU hours on an Intel(R) Xeon(R) CPU E5-4667 server, and all twelve algorithms finished experimentation within the 24 hours time frame. 

Concerning the number of repetitions, we note that with 11 runs we already get a good understanding of the key differences between the algorithms. 11 runs can be enough to get statistical significance, if the differences in performance are substantial. We refer the interested reader to the tutorial~\cite{Hansen18}, which argues that for a first experiment a small number of experiments can suffice. We also note that \textsc{IOHanalyzer} offers statistical tests, in the form of pairwise Kolmogorov-Smirnov tests. An extension to Bayesian inference statistics is in progress. We do not report on these statistical tests here, but refer the interested reader to our work~\cite{CalvoSCD0BL19}, which discusses statistical significance of the results presented here in this work using the framework previously presented in~\cite{CalvoCL18}. We also note that other statistical tests could make a lot of sense to analyze our data, but this is beyond the scope of this work.

\subsection{Performance Measures}
\label{sec:measures}

We are  mostly interested in fixed-target results, i.e., we consider the average ``time'' (=number of function evaluations) needed by each algorithm to find a solution that is at least as good as a certain threshold value. To be very explicit, Figure~\ref{fig:F1D625ERT} is a example showing fixed-target curves, which plot the average number of function evaluations ($y$-axis) needed to find a solution $x$ satisfying $f(x) \ge \phi$, where the target value $\phi$ is the value on the $x$-axis.

Given performance data of $r$ independent runs of an algorithm $A$ with a maximal budget of $B$ function evaluations, the expected running time (ERT) value of $A$ for a target value $v$ is $\tfrac{r-s}{s}B+\text{AHT}$, where $s \le r$ is the number of \emph{successful} runs in which a solution of fitness at least $v$ has been found, and AHT is the average first hitting time of these successful runs. We mostly concentrate on ERT values in our analysis. 

Another important concept in the analysis of IOHs are empirical cumulative distribution function (ECDF) curves, which allow to aggregate performance across different functions. For each of the $d$ different problems $i=1,\ldots,d$ a list of target values $\phi_{i,j}$, $j=1,\ldots,m(i)$, is chosen. The ECDF value at budget $t$ is defined as the fraction of (run,target)-pairs $(s,\phi_{i,j})$ which satisfy that in run $s$ on problem $i$ a solution has been identified that is at least as good as $\phi_{i,j}$. See~\cite{HansenABTT16} for more information and motivation.

\subsection{Function-wise Raw Observations Across Dimensions}
Figures~\ref{fig:ERT-625} and~\ref{fig:ERT-64} depict the ERT of the baseline algorithms on the 625-dimensional and the 64-dimensional functions, respectively, when considering the best function value found by any of the algorithms in any of the runs. These target values are summarized in Table~\ref{tab:targets}.

\begin{table}[t]
\scriptsize
\centering
\begin{tabular}{r||r|r|r|r|r|r|r|r|r|r|r|r|r|r|r|r|r|r|r|r|r|r|r|}
\hline
& $F1$ & $F2$ & $F3$ & $F4$ & $F5$ & $F6$ & $F7$ & $F8$ & $F9$ & $F10$ & $F11$ \\\hline
$n=64$ & 64 & 64 & 2080 & 32 & 57 & 21 & 64 & 33 & 64 & 63.2 & 32  \\ \hline
$n=625$ & 625 & 625 & 195\,625 & 312 & 562 & 208 & 576.4 & 314 & 625 & 625 & 312  \\ \hline
\end{tabular}
\begin{tabular}{r||r|r|r|r|r|r|r|r|r|r|r|r|r|r|r|r|r|r|r|r|r|r|r|}
\hline
& $F12$ & $F13$ & $F14$ & $F15$ & $F16$ & $F17$ & $F18$ & $F19$ & $F20$ & $F21$ & $F22$ & $F23$ \\\hline
$n=64$ & 57 & 21 & 43.8 & 33 & 64 & 64 & 3.981492 & 128 & 230.4 & 384 & 28 & 8  \\ \hline
$n=625$ & 562 & 208 & 36.6 & 314 & 625 & 625 & 4.2655266 & 1\,242 & 2\,420 & 3\,532.8 & 268.4 & 24 \\ \hline
\end{tabular}
\caption{Target values for which the ERT curves in Figures~\ref{fig:ERT-625} and~\ref{fig:ERT-64} are computed.}
\label{tab:targets}
\end{table}

We summarize a few basic observations for each function. 
\begin{enumerate}[F1]
    \item This baseline \onemax problem is easily solved, having the gHC winning (it solves each $n$-dimensional \onemax instance in at most $n+1$ queries), the majority of the algorithms clustered with a practically-equivalent performance, the (1+10)-EA$_{r/2,2r}$ lagging behind, and the vGA outperformed by far. 
	All algorithms locate the global optimum eventually.
    Figure~\ref{fig:F1D625ERT} presents the average fixed-target performance of the algorithms on F1 at $n=625$, in terms of ERT. 
	Evidently, the vGA and the UMDA obtain a clear advantage in the beginning of the optimization process, although the vGA eventually uses the largest number of evaluations, by far, to locate the optimum.
We also see here that, as expected, the performances of the unbiased algorithms (i.e., all algorithms except the vGA) are identical for the 11 runs on instance 1 and the 1 run on 11 different instances. For the vGA this is clearly not the case, the fixed-target performances of these two settings differ substantially.	
\begin{figure}
  \centering
   \includegraphics[width=.9\textwidth]{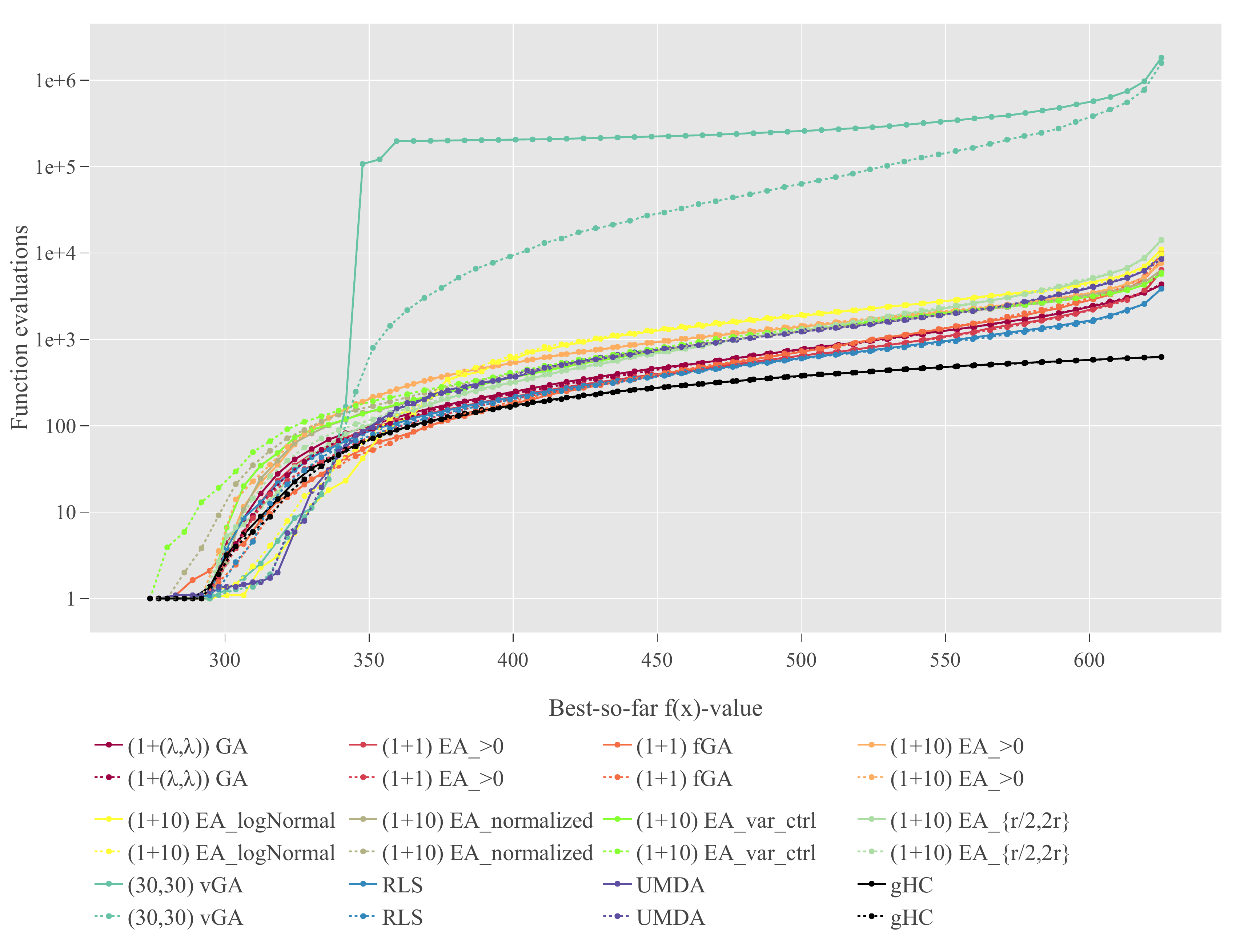}
  \caption{ERT values for F1 (\onemax) at dimension $n=625$ in a fixed-target perspective. The dashed lines are the average running times of 11 independent runs on instance 1, while the solid lines are average running times for one run on each of the eleven different instances 1-6 and 51-55. 
	\label{fig:F1D625ERT}}
\end{figure}
    \item The \leadingones problem introduces more difficulty when compared to F1, with the ERT consistently shifting upward, but it is still easily solved. The gHC wins, the vGA loses, and the majority of the algorithms are again clustered, but now the $(1+(\lambda,\lambda))$-GA lags behind. 
		The UMDA fails to find the optimum within the given time budget, for all tested dimensions except for $n=16$.
    An example of the evolution of the parameter $\lambda$ in the \ga is visualized in Figure~\ref{fig:F2D625MUT_LAMBDA}. We observe -- as expected -- that larger function values are evidently correlated with larger population sizes (and, thus, larger mutation rates).
    \begin{figure}
        \centering
              \includegraphics[width=.6\linewidth]{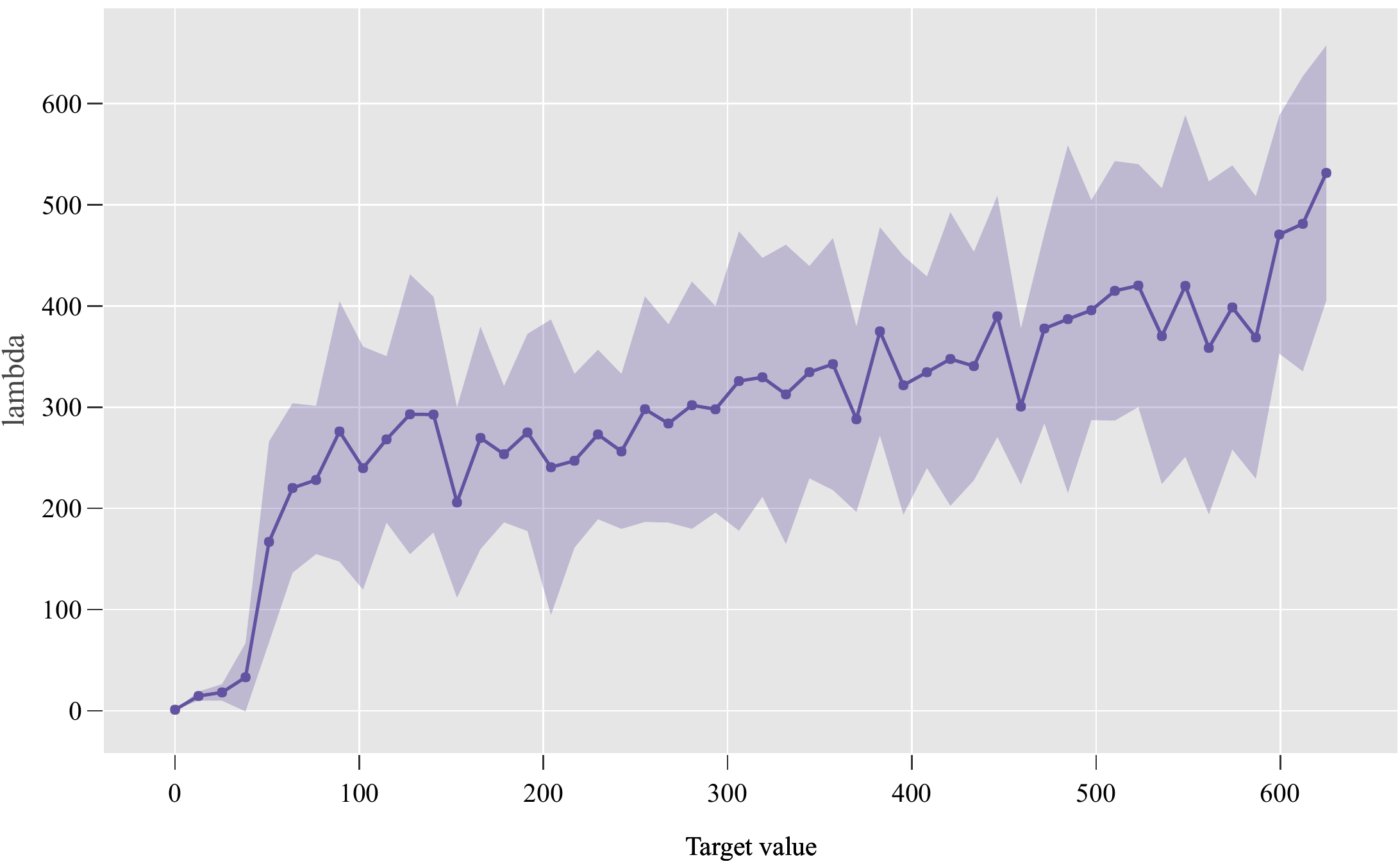}
        \caption{Evolution of the population size $\lambda$ of the \ga on the \leadingones problem F2 at dimension n = 625, correlated to the best-so-far objective function values (horizontal axis). The line shows the average value of $\lambda$ for iterations starting with a best-so-far solution of the value indicated by the $x$-axis. The shade represents the standard deviation.
        \label{fig:F2D625MUT_LAMBDA}}
    \end{figure}
    \item The behavior on this problem, the linear function with harmonic weights, is similar to F1 for most algorithms. Exceptions are the vGA, for which it is slightly easier, and the UMDA, which shows worse performance on F3 than on F1. 
    \item This problem, \onemax with 50\% dummy variables, is the most easily-solved problem in the suite, with an even simpler performance classification -- the gHC performs at the top, the vGA at the bottom, and the rest are tightly clustered. 
	Given the consistent correlation with the F1 performance profiles, across all twelve algorithms, it seems debatable whether or not to keep this function in a benchmark suite, since it seems to offer only limited additional insights, which could be of a rather specialized interest, e.g., for theoretical investigations addressing precise running times of the algorithms.
    \item Solving this problem, \onemax with 10\% dummy variables, exhibits equivalent behavior to F1. Similarly to F4, we suggest to ignore this setup for future benchmarking activities. 
	Note, however, that the exclusion of F4 and F5 does \emph{not} imply that the dummy variables do not play an interesting role -- in an ongoing evaluation of the W-model, we are currently investigating their impact when combined with other W-model transformations. 
    \item The neutrality (``majority vote'') transformation apparently introduces difficulty to the \ga, which exhibits deteriorated performance compared to F1. The vGA, despite a slightly better performance compared to F1, is the worst among the twelve algorithms. 
	At the same time, the (1+10)-EA$_{\lognormal}$ lags behind its competitors in the beginning, but it eventually shows a competitive result in the later optimization process, ending up with an overall fine ERT value. The gHC outperforms all other algorithms also on this function. 
    \item The introduction of local permutations to \onemax, within the current problem, introduced difficulties to all the algorithms.
    The ability to locate the global optimum within the designated budget deteriorated for all of them, except for the (1+10)-EA$_{r/2,2r}$ on ``low-dimensional'' scales ($n\in \left\{16,64,100 \right\}$).  
    Figure \ref{fig:F7D625ERT} depicts the ERT values of the algorithms on F7 at $n=625$, where it is evident that they all failed to locate the global optimum.
	Note that this figure encompasses results for both instantiations (a single instance or 11 instances). 
	The twelve algorithms' performances are clustered in two groups that are associated with two fitness regions (and likely two basins of attraction) - the first around an objective function value of 500 (including the UMDA, with the gHC being the fastest to approach it and get stuck), and the other below 600. It seems that the latter cluster could use additional budget to further improve the results.
	
\begin{figure}
  \centering
   \includegraphics[width=.9\textwidth]{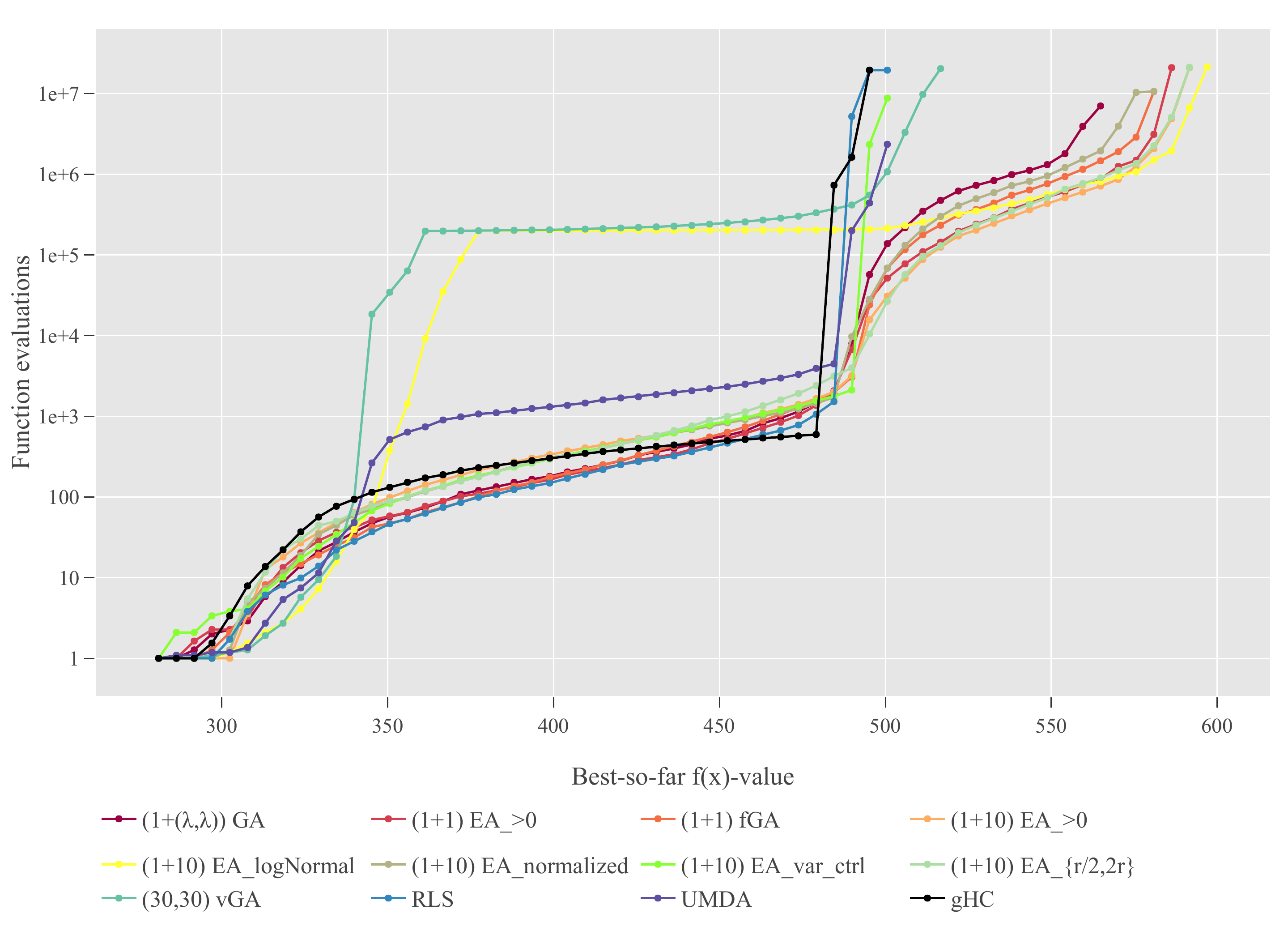}
  \caption{ERT values for F7 at dimension $n=625$ in a fixed-target perspective.  \label{fig:F7D625ERT}}
\end{figure}

\item Being \onemax with the small fitness plateaus induced by the ruggedness function $r_1$, the UMDA performs best on this problem, with the \ga following very closely.
It seems to introduce medium difficulty to all the algorithms, except for the gHC, whose performance is dramatically hampered and becomes worse than the vGA. Interestingly, the ERT values are distributed sparsely compared to other \onemax variants.

\item The UMDA also performs best for this problem, with the (1+10) EA$_{\text{var.}}$ being the runner-up. 
Generally, the behavior on this problem, \onemax with small fitness perturbations, is close to F8, but with certain differences. 
It is evidently harder, as the algorithms experience larger ERT values. 
Importantly, unlike F8, the RLS always fails on F9 (since it gets stuck in local optima), and ``joins'' the gHC and vGA at the bottom of the performance table.  
The \ga also shows worse performance on F9 than on F8.

\item This problem, \onemax with fitness perturbations of block size five, presents a dramatic difficulty to all the algorithms, including the UMDA, which, however, clearly outperforms all other algorithms.  
It is evidently the hardest \onemax variant for all the tested algorithms, among the eight variants studied in this work.
For $n=625$ the UMDA finds the optimum after an average of 141\,243 evaluations, while none of the other algorithms finds a solution better than 575.

\item The gHC performs strongly on this problem, namely the \leadingones with 50\% dummy variables, consistently with its winning behavior on F2. The vGA performs poorly, and the UMDA is also at the bottom of the table. 
Notably, the problem should become easier compared to F2, since the effective number of variables is reduced. The RLS, however, which generally performs well on \leadingones, only ranks third from the bottom on ERT values when solving this problem.

\item  The behavior of the algorithms on this problem, \leadingones with 10\% dummy variables, is very similar to F11, with excellent performance of the gHC. 
However, one major difference is the dramatic deterioration of UMDA and vGA, which fail to find the optimum with given time budget for $n=625$ (see, Figure \ref{fig:ERT-625}). UMDA performs better than vGA for $n=64$, but still obtains clear disadvantage comparing to other algorithms (see Figure \ref{fig:ERT-64}).

\item The introduction of neutrality on \leadingones makes this problem easier in practice (that is, by observing ERT decrease compared to F2). The gHC wins, while the vGA, \ga as well as the UMDA lag behind the other methods. The poor performance of these three algorithms is consistent with their performance on \leadingones. 

\item Being \leadingones with epistasis, this problem introduces high difficulty. 
For high dimensions, $n \in \{64,100,625\}$, none of the algorithms was capable of locating an optimal solution within the allocated budget. 
The vGA tops the ERT values on this problem, followed by the (1+1)-fGA, the (1+10)-EA$_{r/2,2r}$, and the (1+10)-EA$_{\norm}$. On the other hand, three algorithms, namely the gHC, the UMDA, and the RLS, seem to get trapped with low objective function values.

\item The introduction of fitness perturbations to \leadingones makes this problem difficult. 
The UMDA exhibits the worst performance among the competing methods.
The remainder of the algorithms, except for the vGA and the $(1+(\lambda,\lambda))$~GA, are still able to hit the optimum of this problem, but with significantly larger ERT values. The gHC performs best, and the first runner-up is the RLS.

\item The obtained ranks of algorithms, with respect to the ERT values, are similar to those of F15, but generally exhibit higher ERT values. Notably, the UMDA is still the worst performer.

\item As expected, the rugged \leadingones function is the second-hardest among the \leadingones variants, following F14. Only the RLS and the (1+1)-EA are able to hit the optimum in dimension 625, while the gHC has a diminished performance on this problem. This can be explained by the fact that the gHC has a very high probability of getting stuck in a local traps, while the RLS is capable of performing random walks about local optima, until eventually escaping them (e.g., by flipping the right bit when all the consecutive four bits are also identical to the target string). This is of course a rare event, and the ERT values are therefore significantly worse than all other \leadingones variants, except F14. As on the previous two functions, the UMDA performs poorly, with similar ERT values as the gHC. 
 
Comparing to F10, the effect of the fitness permutation $r_3$ on \leadingones is not as significant as on \onemax, which can be explained by the ability of most of the algorithms to perform random walks about local traps, through which the four first bits of the tail are eventually set correctly, at which point flipping the significant bit (i.e., the bit in position $\LO(x)+1$) results in a \leadingones fitness increase of at least five, and consequently a fitness increase of at least one for the problem $r_3 \circ \leadingones$. This candidate solution is thus accepted by all of our algorithms, and the next phase of optimizing the following consecutive five bits begins.    
 
\item The LABS problem is the most complex problem in our assessment. For the higher dimensions, $n \in \{64,100,625\}$, none of the algorithms obtained the maximally attainable values, or got fitness values close to those of the best-known sequences (see, e.g., \cite{LABS_Packebusch2016}). Additionally, a couple of algorithms (e.g., the gHC and the RLS) did not succeed to escape low-quality ``local traps'' on most dimensions. 
Surprisingly, the vGA was superior to the other algorithms at $n=16$ but, as expected, over the higher dimensions presented weaker performance. 
Notably, the UMDA outperforms the other methods at $n=625$.
 
\item The simplest problem among the Ising instances. Most of the algorithms exhibited similar performance, except for the vGA, the UMDA and the gHC, which obtained weak results. The latter preformed worst among all algorithms, and obtained the lowest objective function values across all the dimensions for the given time budget. As a demonstration of the performance statistics that \tool provides, average fixed-target and fixed-budget running times are provided in Figure~\ref{fig:F19_demo}. This figure illustrates that ERT values tell only one side of the story: the performance of UMDA is comparable to that of the other algorithms for all targets up to around 170; only then it starts to perform considerably worse.  

\item In contrast to its poor performance on the 1D-Ising (F19), the gHC outperformed the other algorithms on the 2D-Ising for target values up to around 2,300 ($d=625$), after which its performance becomes worse than most of the other algorithms, except for the vGA, which is consistently the worst except for a few initial target values. For the 16-dimensional F20 problem, gHC performs best for all recorded target values. For $d=625$, however, the (1+10) EA achieves the best ERT value for the target recorded in Table~\ref{tab:targets}, followed by the (1+1) EA, the (1+10) EA$_{\vars}$, and the fGA.  

\item As expected, the most complex among the Ising model instances. The observed performances resemble the observations on F20, whereas the vGA was among the slowest and the gHC was among the fastest. In addition, the RLS preformed poorly and obtained worse ERT values compared to the other algorithms (apart from the vGA, which performs even worse). The UMDA improved its ranking compared to F19, exhibiting performance close to the vGA. For $d=625$, the best ERT is obtained by the (1+10) EA$_{\lognormal}$.  

\item None of the algorithms succeeded in locating the global optimum across all dimensions of this problem. This is explained by the existence of a local optimum with a strong basin of attraction. The gHC and the vGA exhibited inferior performance compared to the other algorithms.

\item Some algorithms failed to locate the global optimum of the N-Queens problem in high dimensions, yet the vGA, the gHC and the UMDA constantly possessed the worst ERT values. 
Fine performance was observed for the (1+10)-EA$_{>0}$ and the (1+10) EA$_{\lognormal}$.
\end{enumerate}

\subsection{Grouping of Functions and Algorithms}
\label{sec:ERT}


In this section we are aiming to recognize patterns and identify classes within (i) the set of all functions, and (ii) the set of all algorithms.
Our analyses are based on human experts' observations, alongside automated clustering of the mean ERT vectors using $K$-\textit{means} \cite{hastie2013elements}. 

\paragraph{Functions' Empirical Grouping}

It is evident that problems F1-F6, F11-F13 and F15-F16 are  treated relatively easily by the majority of the algorithms, with those functions based on \leadingones (i.e., F2, F11-13, F15, F16) 
being more challenging within this group.
On the other extreme, F7, F9-F10, F14, F18-F19 and F22 evidently constitute a class of hard problems, on which all algorithms consistently exhibit difficulties (except for $n=16$); the LABS function (F18) seems the most difficult among them.
F8, the instances of the Ising model (F19-F21), as well as the NQP (F23), constitute a class of moderate level of difficulty.

\paragraph{Algorithms' Observed Trends}
The gHC and the vGA usually exhibited extreme performance with respect to the other algorithms. The vGA consistently suffers from poor performance over all functions, while the gHC either leads the performance on certain functions or performs very poorly on other. 
The gHC's behavior is to be expected, since it is correlated with the existence of local traps (by construction) -- for instance, it consistently performs very well on F1-F6, while having difficulties on F7-F10. Clearly, RLS also gets trapped by the deceptive functions, while at the same time it shows fine performance on most of the non-deceptive problems. 
The UMDA's performance stands out, and also appears as an independent cluster in the automated K-Means analysis. Evidently, it performs well on the \onemax-based problems, but fails to optimize the \leadingones function F2 and its derivatives F11-F17, with the exception of F11 and F13 -- a behavior that might be interesting to analyze further in future work. 
Otherwise, we observe one primary class of algorithms exhibiting equivalent performance over all problems in all dimensions:
The 7 algorithms $(1+(\lambda,\lambda))$-GA, (1+1)-EA, (1+10)-EA$_{\vars}$, (1+10)-EA, (1+10)-EA$_{\norm}$, (1+10)-EA$_{r/2,2r}$, 
and (1+1)-fGA behave consistently, typically exhibiting fine performance. In terms of ERT values, the (1+10)-EA$_{\lognormal}$ could also be grouped into this class of seven algorithms, but it behaves quite differently during the optimization process, often showing an opposite trend of convergence speed at the early stages of the optimization procedure. 



\begin{figure}
              \centering
              \includegraphics[width=1\columnwidth]{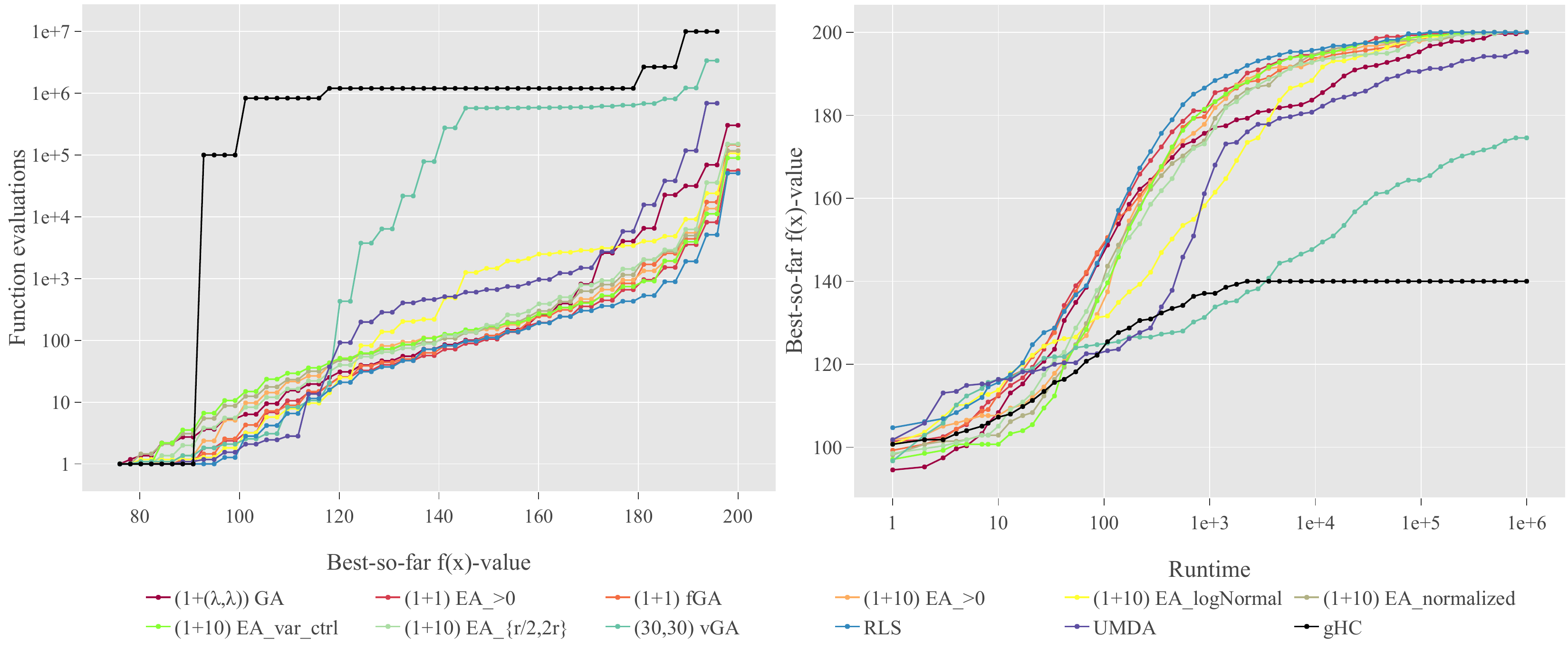}
            \caption{Demonstration of the basic performance plots for F19 at dimension $n=100$: [LEFT] best obtained values as a function of evaluations calls (``fixed-target perspective''), versus [RIGHT] evaluations calls as a function of best obtained values (''fixed-budget perspective''). For F19, these patterns of relative behavior are observed across all dimensions.\label{fig:F19_demo}}
\end{figure}

\paragraph{Ranking}
We also examined the overall number of runs per test-function in which an algorithm successfully located the best recorded value -- the so-called \textit{hitting number}. 
We then grouped those hitting numbers by dimension, and ranked the algorithms per each dimension. The (1+10)-EA$_{r/2,2r}$ consistently leads the grouped hitting numbers on the ``low-dimensional'' functions ($n \in \left\{16,~64 \right\}$), with (1+1)-fGA and (1+10)-EA$_{\norm}$ being together the first runner-up. The (1+10)-EA also exhibits high ranking across all dimensions.
(1+10)-EA$_{\norm}$ leads the grouped hitting numbers on $n=100$, whereas the (1+1)-EA leads the hitting numbers on the ``high-dimensional'' functions at $n=625$, with (1+10)-EA being the runner-up.
Across all dimensions, UMDA, gHC and vGA are with the lowest rankings.

\paragraph{Visual Analytics}
As a demonstration of the performance statistics offered by \tool, we provide snapshots of visual analytics that supported our examination. 
Figure~\ref{fig:F19_demo} depicts basic performance plots for F19 at dimension $n=100$, in so-called \textit{fixed-target} and \textit{fixed-budget} perspectives. For clarity of the plots we only show the ERT values and the average function values achieved per each budget, respectively. Standard deviations as well as the $2,5,10,15,50,75,90,95,98\%$ quantiles are available on \url{http://iohprofiler.liacs.nl/}.  

In Figure~\ref{fig:ECDF_easy625} we provide two plots obtained from our new module which computes ECDF curves for user-specified target values. The plot on the left depicts an ECDF curve for the ``easily-solved'' functions identified above (i.e., F1-F6, F11-F13, and F15-F16) in dimension $n=625$. The one on the right shows the ECDF curves across all 23 benchmark functions. For both figures we have chosen ten equally spaced target values per each function, with the largest value being again the best function value identified by any of the algorithms in any of the runs. Since the number of ``easy'' problems dominates our overall assessment the curves on the right are to a large extent dominated by the performances depicted on the left. This indicates once again the need for a thorough revision of our benchmark selection. 

\begin{figure}[t]
        \centering
        \begin{subfigure}{.5\textwidth}
              \centering
              \includegraphics[width=0.95\columnwidth]{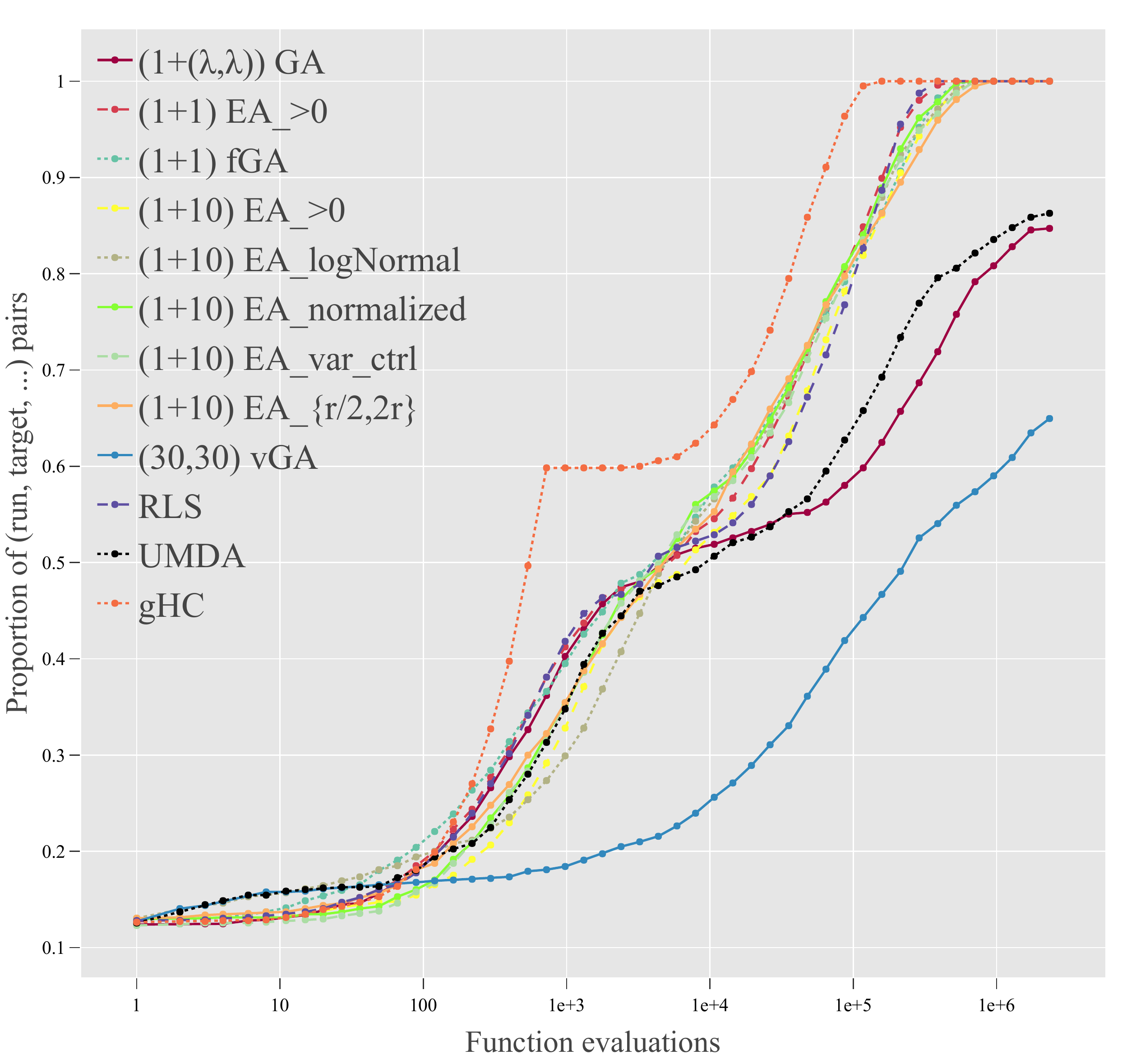}
        \end{subfigure}%
        \begin{subfigure}{.5\textwidth}
              \centering
              \includegraphics[width=0.95\columnwidth]{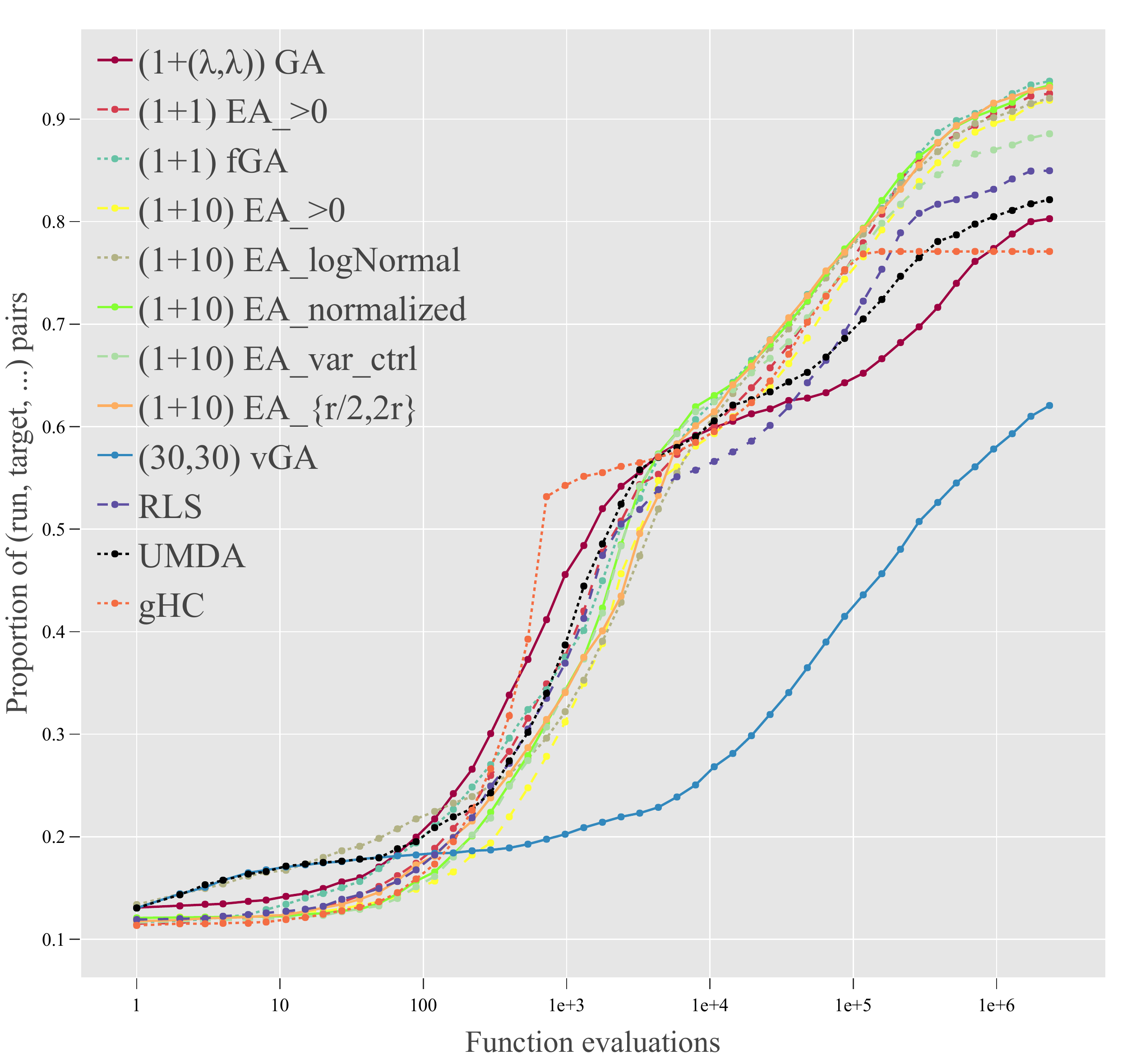}
        \end{subfigure}
            \caption{ECDF curve for the class of ``easily-solved'' functions in dimension $n=625$: F1-F6, F11-F13, and F15-F16 [LEFT] and of all 23 functions [RIGHT], with respect to equally spaced target values.\label{fig:ECDF_easy625}}
\end{figure}

\subsection{Unbiasedness}\label{sec:biasedness}
Following our experimental planning to test the hypothesized ``biasedness'' effect for the vGA, we compared its averaged performance on instance 1 versus on all the other instances (1-6 and 51-55) altogether.
Figure~\ref{fig:F1F2_bias_boxplots} depicts a comparison of attained objective function values, by means of box-plots, on F1 and on F2 for $n=64$.
Performance deterioration is indeed evident on the permuted instances; that is, instances 51-55, for which the base functions are composed with a $\sigma$-transformation of the bit strings, as described in Section~\ref{sec:unbiasedness}. 
The box-plots in Figure~\ref{fig:F1F2_bias_boxplots} show very clearly that the vGA treats the plain F1 and F2 much better, in terms of attained target values, than their transformed variants. The plots are for $n=64$ and after exhausting the full budget of $100n^2$ function evaluations.


\begin{figure}
        \centering
        \begin{subfigure}{.5\textwidth}
              \centering
              \includegraphics[width=\columnwidth, trim=0mm 0mm 10mm 0mm, clip]{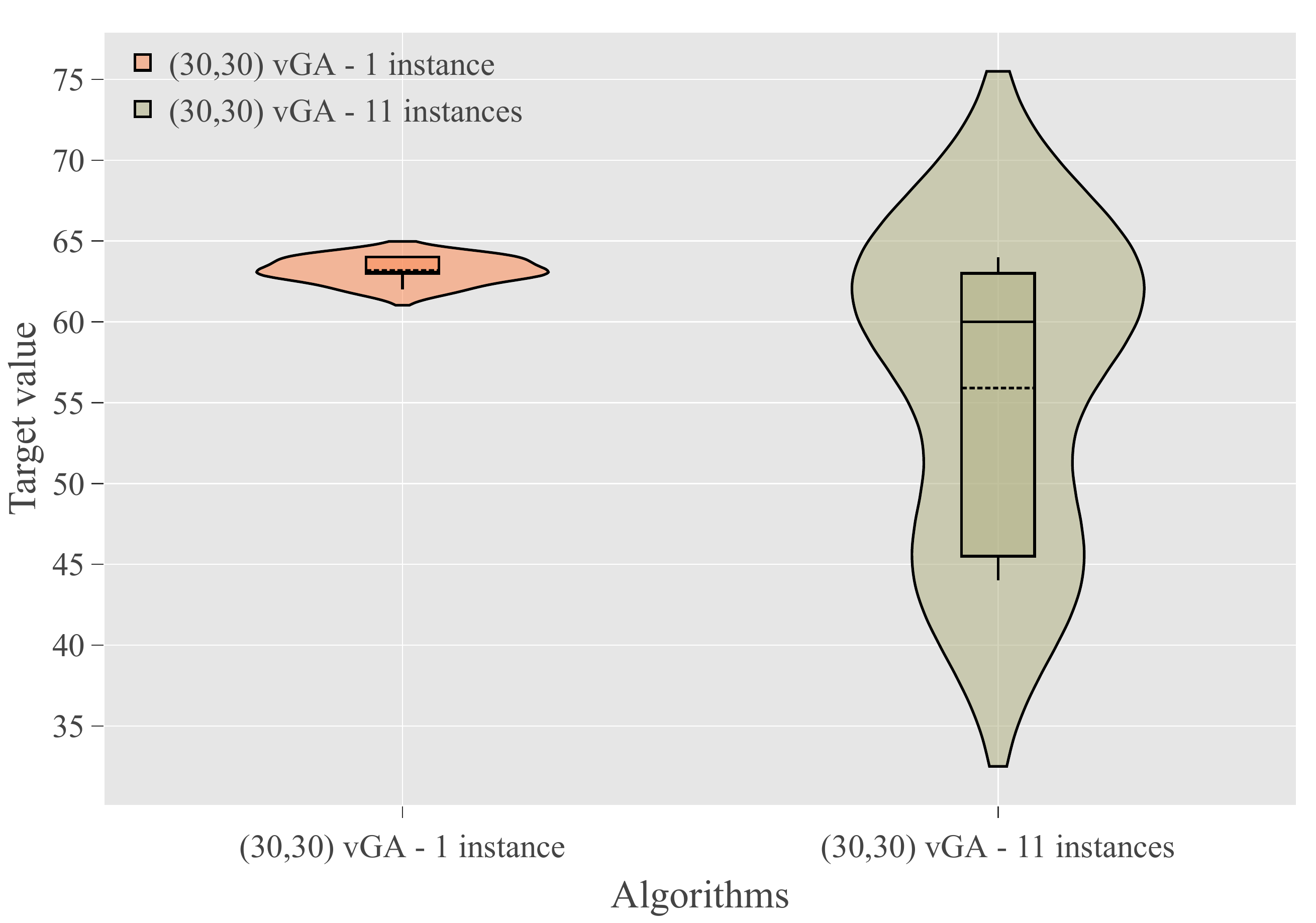}
        \end{subfigure}%
        \begin{subfigure}{.5\textwidth}
              \centering
              \includegraphics[width=\columnwidth, trim=0mm 0mm 10mm 0mm, clip]{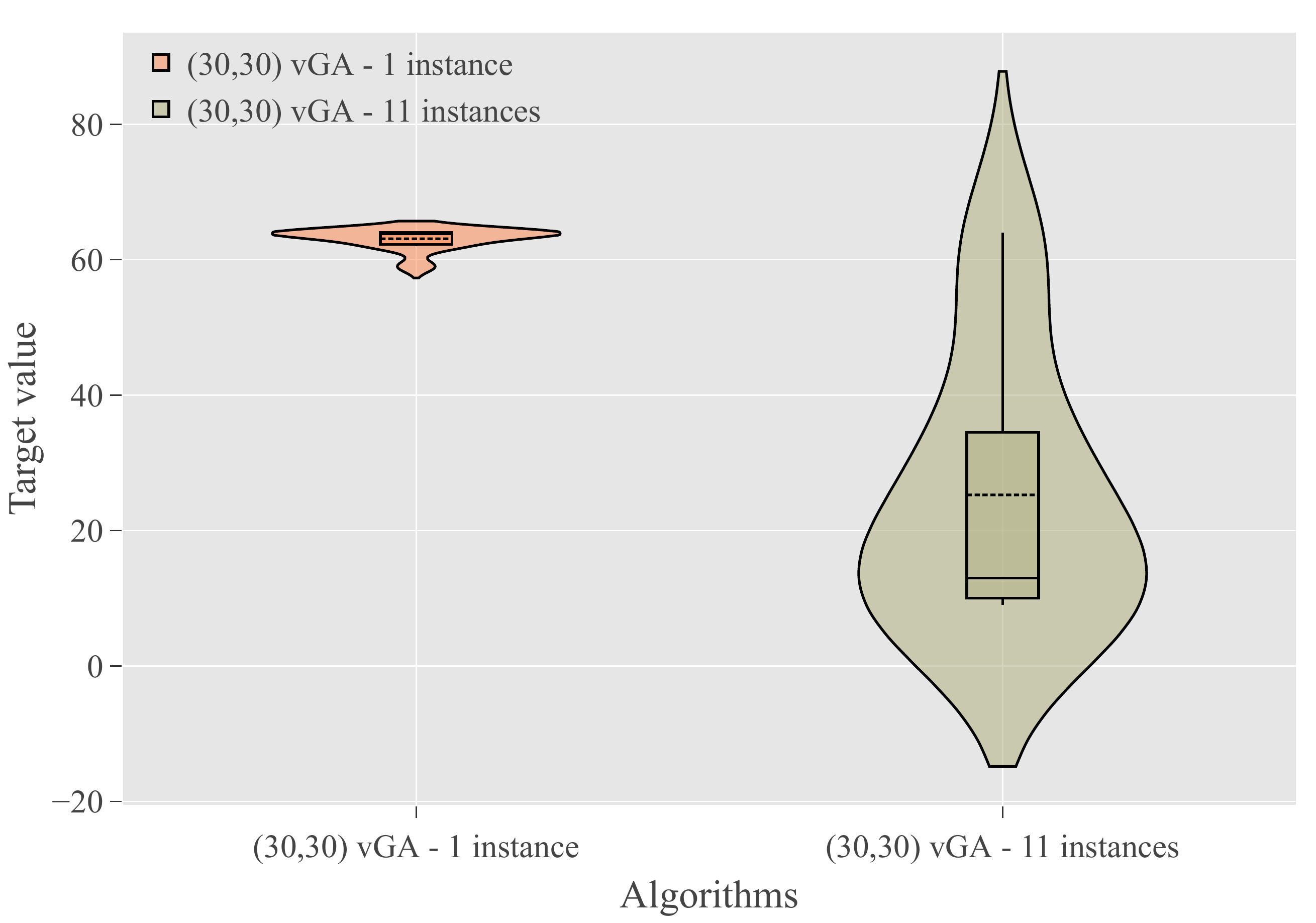}
        \end{subfigure}
        \caption{Statistical box-plots for vGA's attained function values on instance 1 alone versus on the eleven different instances 1-6 and 51-55, after exploiting the entire budget (namely, 409\,600 function evaluations): F1 [LEFT] and F2 [RIGHT]. Both plots are for $n=64$. \label{fig:F1F2_bias_boxplots}}
\end{figure}
    

\subsection{Aftermath}\label{sec:discussion} 

The reported experiments reveal interesting findings on the test-functions and the algorithms' behavior when solving them -- among which we highlight a few.
By construction, the current test-suite proposed functions with various degrees of difficulty. 
Certain functions are inherently difficult (e.g., F18), and some synthetically (e.g., F7 and F14, regenerated by ``epistasis''). 
Our empirical observations on such synthetically regenerated hard problems corroborated the effectiveness of the W-model in such a benchmarking environment. However, we have also observed that some of the components of the W-model, in particular the introduction of dummy variables, might be less interesting as ``stand-alone''-effects. A revised selection of the W-model functions is currently under investigation. 

Finally, we note that the hardness of all functions consistently increase with their problem dimension -- exhibiting a desirable property that we mentioned in Section~\ref{sec:mindset}.

\section{Outlook}
\label{sec:outlook}

Among the many possible directions for future work, we consider the following ones particularly interesting. 

\textbf{Additional Performance Measures:} While \tool already provides a very detailed assessment of algorithms' performance data, we are continuously strengthening its statistical repertoire by introducing new performance measures any by devising better procedures. We have implemented for this report and for future use in \tool the possibility to generate ECDF curves, for a user-defined set of functions and target values, thereby following the interactive performance evaluation paradigm which distinguished \tool from other existing benchmarking platforms (where the targets or budgets are typically set fixed). Going forward, we suggest to include modules that allow performance comparisons across different dimensions. To shed better light on the tradeoff with respect to time, quality, and robustness, we are also considering to add an automated computation of the empirical attainment function, as available from~\cite{eaf}. We are furthermore looking into the option of adding a variant of the so-called performance profiles~\cite{more2009benchmarking} for comparison across multiple dimensions.
In terms of statistical tests, two-sample nonparametric tests (e.g., Mann–Whitney U or Kolmogorov–Smirnov test) can be applied pairwise among algorithms with corrections (e.g., Holm-Bonferroni correction). The Kolmogorov-Smirnov tests are available in \textsc{IOHanalyzer}, while extension, in particular in terms of nonparametric tests that are designed for two or more samples, e.g., Kruskal–Wallis or Friedman test, are left for future work. For the large-scale multiple testing scenario (thousands of pairwise tests are performed, for instance), we are considering to add Bayesian inference, as suggested, e.g., in~\cite{BenavoliCDZ17}. Such an approach has recently been proposed for comparing performance of IOHs~\cite{CalvoCL18} and has been applied to the data set of this paper in~\cite{CalvoSCD0BL19}. 

\textbf{Feature-Based Analyses:} Another interesting extension of \textsc{IOHanalyzer} would be the design of modules that allow us to couple the performance evaluation with an analysis of the fitness landscape of the considered problems. Such \emph{feature-based analyses} are at the heart of algorithm selection techniques~\cite{kerschke2018survey}, which use landscape features and performance data to build a model that predicts how the tested algorithms will perform on a previously \textit{unseen} problem. Similar approaches can be found in per-instance-algorithm configuration (PIAC) approaches, which have recently shown very promising performance in the context of continuous black-box optimization~\cite{BelkhirDSS17}. A key step towards such a feature-based performance analysis are the selection and the efficient computation of meaningful features. While in continuous optimization a large set of features has been defined and can be computed with the flacco package~\cite{flacco}, the research community currently lacks a meaningful analog for discrete optimization problems. We note though, that several advances in this direction have been made, including the above-introduced features covered by the W-model (size of the effective dimension, neutrality, epistasis, ruggedness) and the local optima networks (see~\cite{ThomsonVOVM18,ThomsonVOVC18} and references mentioned therein). 
We suggest to start a first prototype using these existing features, while at the same time intensifying research efforts to find additional landscape features that can be used to characterize pseudo-Boolean optimization problems.  

\textbf{Critical Assessment of Benchmark Problems:} With such a feature-based approach, we also hope to develop a more systematic approach towards the identification of problem characteristics that are not well represented in the selection of problems described above. We emphasize once again the fact that we see the here-suggested set of benchmark problems as a \emph{first step} towards a sound benchmarking suite, not as a static ``ultimate'' selection. Quite the contrary, another important direction of our future research concerns the identification, critical selection, and implementation of additional benchmark problems. We believe that a good benchmark environment should be dynamic, with the possibility to add new problems, so that users can focus their experimentation on problems relevant to their work. \tool is built with this functionality in mind, making it a particularly suitable testbed for our studies.   

\textbf{Combinations of W-model Transformations:} 
As discussed in Section~\ref{sec:W}, the transformations of the W-model can be combined with each other. To analyze the individual effects of each transformation, and in order to keep the size of the experimental setup reasonable, we have not considered such combinations in this work. A critical consideration of adding such combinations, and of extending the base transformations (e.g., with respect to the fitness transformation, but also the size of the neutrality transformation, etc.) forms another research line that we are currently addressing in a parallel work stream.

\textbf{Integration of Algorithm Design Software:} IOHs are to a large extent modular algorithms, whose components can be exchanged and executed in various different ways. This has let the community to develop software which enables an easier algorithmic design. Examples for such software are ParadisEO~\cite{paradiseo} for single-objective and multi-objective optimization and jMetal~\cite{jMetal} for multi-objective algorithms. Building or integrating such software could allow much more comprehensive algorithm benchmarking, and could eventually automate the detection of promising algorithmic variants. In the continuous domains, the modular CMA-ES framework~\cite{modularCMAES} can be seen as a proof of concept for this idea: it was shown there that some of the automatically generated CMA-ES variants could improve upon existing human-designed algorithms. In a similar direction, we are also planning to ease parallelization of \textsc{IOHexperimenter}, so that parametric algorithms can be batch-tested for various configurations.

\section*{Acknowledgments}
We are very grateful to our colleagues Arina Buzdalova and Maxim Buzdalov (both at ITMO University, St.~Petersburg, Russia), Johann Dreo (Thales Research), Michal Horovitz and Mordo Shalom (both at Tel-Hai College, Israel), Dirk Sudholt (Sheffield, UK), and Thomas Weise (Hefei University, China) for valuable discussions around different aspects of benchmarking IOHs. We have very much appreciated the detailed comments of the anonymous reviewers of this paper, which have helped us improve the presentation of our contribution. In particular, the suggestion to add an EDA to the set of baseline algorithms was made by one of the reviewers. 

Our work was supported by the Chinese scholarship council (CSC No. 201706310143), a public grant as part of the Investissement d'avenir project, reference ANR-11-LABX-0056-LMH, LabEx LMH, in a joint call with the Gaspard Monge Program for optimization, operations research, and their interactions with data sciences, 
by Paris Ile-de-France Region, 
and by COST Action CA15140 ``Improving Applicability of Nature-Inspired Optimisation by Joining Theory and Practice (ImAppNIO)''.

}


\begin{thebibliography}{75}
\expandafter\ifx\csname natexlab\endcsname\relax\def\natexlab#1{#1}\fi
\providecommand{\url}[1]{\texttt{#1}}
\providecommand{\href}[2]{#2}
\providecommand{\path}[1]{#1}
\providecommand{\DOIprefix}{doi:}
\providecommand{\ArXivprefix}{arXiv:}
\providecommand{\URLprefix}{URL: }
\providecommand{\Pubmedprefix}{pmid:}
\providecommand{\doi}[1]{\href{http://dx.doi.org/#1}{\path{#1}}}
\providecommand{\Pubmed}[1]{\href{pmid:#1}{\path{#1}}}
\providecommand{\bibinfo}[2]{#2}
\ifx\xfnm\relax \def\xfnm[#1]{\unskip,\space#1}\fi
\bibitem[{Kerschke et~al.(2018)Kerschke, Hoos, Neumann, and
  Trautmann}]{kerschke2018survey}
\bibinfo{author}{P.~Kerschke}, \bibinfo{author}{H.~H. Hoos},
  \bibinfo{author}{F.~Neumann}, \bibinfo{author}{H.~Trautmann},
\newblock \bibinfo{title}{Automated algorithm selection: Survey and
  perspectives},
\newblock \bibinfo{journal}{CoRR} \bibinfo{volume}{abs/1811.11597}
  (\bibinfo{year}{2018}). \URLprefix \url{http://arxiv.org/abs/1811.11597}.
  \href{http://arxiv.org/abs/1811.11597}{{\tt arXiv:1811.11597}}.
\bibitem[{Wasik et~al.(2016)Wasik, Antczak, Badura, Laskowski, and
  Sternal}]{WasikABLS16}
\bibinfo{author}{S.~Wasik}, \bibinfo{author}{M.~Antczak},
  \bibinfo{author}{J.~Badura}, \bibinfo{author}{A.~Laskowski},
  \bibinfo{author}{T.~Sternal},
\newblock \bibinfo{title}{Optil.io: Cloud based platform for solving
  optimization problems using crowdsourcing approach},
\newblock in: \bibinfo{booktitle}{Proc. of {ACM} Conference on Computer
  Supported Cooperative Work and Social Computing (CSCW'16), Companion Volume},
  \bibinfo{publisher}{{ACM}}, \bibinfo{year}{2016}, pp.
  \bibinfo{pages}{433--436}. \URLprefix
  \url{https://doi.org/10.1145/2818052.2869098}.
  \DOIprefix\doi{10.1145/2818052.2869098}.
\bibitem[{Weise(2016)}]{weisebench}
\bibinfo{author}{T.~Weise}, \bibinfo{title}{Optimization benchmarking},
  \bibinfo{year}{2016}. \bibinfo{note}{Available at
  \url{http://optimizationbenchmarking.github.io/}}.
\bibitem[{Rapin and Teytaud(2018)}]{nevergrad}
\bibinfo{author}{J.~Rapin}, \bibinfo{author}{O.~Teytaud},
  \bibinfo{title}{{Nevergrad - A gradient-free optimization platform}},
  \bibinfo{howpublished}{\url{https://GitHub.com/FacebookResearch/Nevergrad}},
  \bibinfo{year}{2018}.
\bibitem[{Hansen et~al.(2010)Hansen, Auger, Ros, Finck, and
  Po\v{s}\'{\i}k}]{BBOB_GECCO2010}
\bibinfo{author}{N.~Hansen}, \bibinfo{author}{A.~Auger},
  \bibinfo{author}{R.~Ros}, \bibinfo{author}{S.~Finck},
  \bibinfo{author}{P.~Po\v{s}\'{\i}k},
\newblock \bibinfo{title}{Comparing results of 31 algorithms from the black-box
  optimization benchmarking bbob-2009},
\newblock in: \bibinfo{booktitle}{Proceedings of the 12th Annual Conference
  Companion on Genetic and Evolutionary Computation}, GECCO '10,
  \bibinfo{publisher}{ACM}, \bibinfo{address}{New York, NY, USA},
  \bibinfo{year}{2010}, pp. \bibinfo{pages}{1689--1696}. \URLprefix
  \url{http://doi.acm.org/10.1145/1830761.1830790}.
  \DOIprefix\doi{10.1145/1830761.1830790}.
\bibitem[{Hansen et~al.(2016)Hansen, Auger, Mersmann, Tu\v{s}ar, and
  Brockhoff}]{COCO16platform}
\bibinfo{author}{N.~Hansen}, \bibinfo{author}{A.~Auger},
  \bibinfo{author}{O.~Mersmann}, \bibinfo{author}{T.~Tu\v{s}ar},
  \bibinfo{author}{D.~Brockhoff},
\newblock \bibinfo{title}{{COCO:} {A} platform for comparing continuous
  optimizers in a black-box setting},
\newblock \bibinfo{journal}{CoRR} \bibinfo{volume}{abs/1603.08785}
  (\bibinfo{year}{2016}). \URLprefix \url{http://arxiv.org/abs/1603.08785}.
  \href{http://arxiv.org/abs/1603.08785}{{\tt arXiv:1603.08785}}.
\bibitem[{Tusar et~al.(2016)Tusar, Brockhoff, Hansen, and Auger}]{TusarBHA16}
\bibinfo{author}{T.~Tusar}, \bibinfo{author}{D.~Brockhoff},
  \bibinfo{author}{N.~Hansen}, \bibinfo{author}{A.~Auger},
\newblock \bibinfo{title}{{COCO:} the bi-objective black box optimization
  benchmarking (bbob-biobj) test suite},
\newblock \bibinfo{journal}{CoRR} \bibinfo{volume}{abs/1604.00359}
  (\bibinfo{year}{2016}). \URLprefix \url{http://arxiv.org/abs/1604.00359}.
\bibitem[{Tusar et~al.(2019)Tusar, Brockhoff, and Hansen}]{TusarBH19}
\bibinfo{author}{T.~Tusar}, \bibinfo{author}{D.~Brockhoff},
  \bibinfo{author}{N.~Hansen},
\newblock \bibinfo{title}{Mixed-integer benchmark problems for single- and
  bi-objective optimization},
\newblock in: \bibinfo{booktitle}{Proc. of Genetic and Evolutionary Computation
  Conference (GECCO'19)}, \bibinfo{publisher}{ACM}, \bibinfo{year}{2019}, pp.
  \bibinfo{pages}{718--726}. \URLprefix
  \url{https://doi.org/10.1145/3321707.3321868}.
  \DOIprefix\doi{10.1145/3321707.3321868}.
\bibitem[{Doerr et~al.(2018)Doerr, Wang, Ye, van Rijn, and
  B{\"a}ck}]{IOHprofiler}
\bibinfo{author}{C.~Doerr}, \bibinfo{author}{H.~Wang}, \bibinfo{author}{F.~Ye},
  \bibinfo{author}{S.~van Rijn}, \bibinfo{author}{T.~B{\"a}ck},
\newblock \bibinfo{title}{{IOHprofiler: A Benchmarking and Profiling Tool for
  Iterative Optimization Heuristics}},
\newblock \bibinfo{journal}{arXiv e-prints:1810.05281}  (\bibinfo{year}{2018}).
  \href{http://arxiv.org/abs/1810.05281}{{\tt arXiv:1810.05281}},
  \bibinfo{note}{available at \url{https://arxiv.org/abs/1810.05281}}.
\bibitem[{Doerr et~al.(2019{\natexlab{a}})Doerr, Ye, Horesh, Wang, Shir, and
  B{\"{a}}ck}]{IOHprofilerGECCO19}
\bibinfo{author}{C.~Doerr}, \bibinfo{author}{F.~Ye},
  \bibinfo{author}{N.~Horesh}, \bibinfo{author}{H.~Wang},
  \bibinfo{author}{O.~M. Shir}, \bibinfo{author}{T.~B{\"{a}}ck},
\newblock \bibinfo{title}{Benchmarking discrete optimization heuristics with
  {IOHprofiler}},
\newblock in: \bibinfo{booktitle}{Companion Material of Proc. of Genetic and
  Evolutionary Computation Conference (GECCO'19)}, \bibinfo{publisher}{ACM},
  \bibinfo{year}{2019}{\natexlab{a}}, pp. \bibinfo{pages}{1798--1806}.
  \URLprefix \url{https://doi.org/10.1145/3319619.3326810}.
  \DOIprefix\doi{10.1145/3319619.3326810}.
\bibitem[{Doerr et~al.(2019{\natexlab{b}})Doerr, Wang, Ye, van Rijn, and
  B{\"a}ck}]{IOhprofiler-github}
\bibinfo{author}{C.~Doerr}, \bibinfo{author}{H.~Wang}, \bibinfo{author}{F.~Ye},
  \bibinfo{author}{S.~van Rijn}, \bibinfo{author}{T.~B{\"a}ck},
  \bibinfo{title}{Github page of {IOHprofiler} data sets},
  \bibinfo{year}{2019}{\natexlab{b}}. \bibinfo{note}{Available at
  \url{https://github.com/IOHprofiler}}.
\bibitem[{Shir et~al.(2018)Shir, Doerr, and B{\"{a}}ck}]{ShirDB18}
\bibinfo{author}{O.~M. Shir}, \bibinfo{author}{C.~Doerr},
  \bibinfo{author}{T.~B{\"{a}}ck},
\newblock \bibinfo{title}{Compiling a benchmarking test-suite for combinatorial
  black-box optimization: a position paper},
\newblock in: \bibinfo{booktitle}{Proc. of Genetic and Evolutionary Computation
  Conference ({GECCO'18}), Companion}, \bibinfo{publisher}{ACM},
  \bibinfo{year}{2018}, pp. \bibinfo{pages}{1753--1760}. \URLprefix
  \url{https://doi.org/10.1145/3205651.3208251}.
  \DOIprefix\doi{10.1145/3205651.3208251}.
\bibitem[{Mersmann et~al.(2011)Mersmann, Bischl, Trautmann, Preuss, Weihs, and
  Rudolph}]{ExploratoryGECCO2011}
\bibinfo{author}{O.~Mersmann}, \bibinfo{author}{B.~Bischl},
  \bibinfo{author}{H.~Trautmann}, \bibinfo{author}{M.~Preuss},
  \bibinfo{author}{C.~Weihs}, \bibinfo{author}{G.~Rudolph},
\newblock \bibinfo{title}{Exploratory landscape analysis},
\newblock in: \bibinfo{booktitle}{Proc. of Genetic and Evolutionary Computation
  (GECCO'11)}, \bibinfo{publisher}{ACM}, \bibinfo{year}{2011}, pp.
  \bibinfo{pages}{829--836}. \URLprefix
  \url{http://doi.acm.org/10.1145/2001576.2001690}.
  \DOIprefix\doi{10.1145/2001576.2001690}.
\bibitem[{Lehre and Witt(2012)}]{LehreW12}
\bibinfo{author}{P.~K. Lehre}, \bibinfo{author}{C.~Witt},
\newblock \bibinfo{title}{Black-box search by unbiased variation},
\newblock \bibinfo{journal}{Algorithmica} \bibinfo{volume}{64}
  (\bibinfo{year}{2012}) \bibinfo{pages}{623--642}.
\bibitem[{Dubhashi and Panconesi(2009)}]{xxxDubhashiP98}
\bibinfo{author}{D.~P. Dubhashi}, \bibinfo{author}{A.~Panconesi},
  \bibinfo{title}{Concentration of Measure for the Analysis of Randomised
  Algorithms}, \bibinfo{publisher}{Cambridge University Press},
  \bibinfo{year}{2009}.
\bibitem[{Garnier et~al.(1999)Garnier, Kallel, and Schoenauer}]{GarnierKS99}
\bibinfo{author}{J.~Garnier}, \bibinfo{author}{L.~Kallel},
  \bibinfo{author}{M.~Schoenauer},
\newblock \bibinfo{title}{Rigorous hitting times for binary mutations},
\newblock \bibinfo{journal}{Evolutionary Computation} \bibinfo{volume}{7}
  (\bibinfo{year}{1999}) \bibinfo{pages}{173--203}.
\bibitem[{Doerr et~al.(2015)Doerr, Doerr, and Ebel}]{DoerrDE15}
\bibinfo{author}{B.~Doerr}, \bibinfo{author}{C.~Doerr},
  \bibinfo{author}{F.~Ebel},
\newblock \bibinfo{title}{From black-box complexity to designing new genetic
  algorithms},
\newblock \bibinfo{journal}{Theoretical Computer Science} \bibinfo{volume}{567}
  (\bibinfo{year}{2015}) \bibinfo{pages}{87--104}.
\bibitem[{Doerr and Doerr(2018)}]{DoerrD18ga}
\bibinfo{author}{B.~Doerr}, \bibinfo{author}{C.~Doerr},
\newblock \bibinfo{title}{Optimal static and self-adjusting parameter choices
  for the $(1+(\lambda,\lambda))$ genetic algorithm},
\newblock \bibinfo{journal}{Algorithmica} \bibinfo{volume}{80}
  (\bibinfo{year}{2018}) \bibinfo{pages}{1658--1709}.
\bibitem[{Erd{\H{o}}s and R{\'e}nyi(1963)}]{ErdR63}
\bibinfo{author}{P.~Erd{\H{o}}s}, \bibinfo{author}{A.~R{\'e}nyi},
\newblock \bibinfo{title}{On two problems of information theory},
\newblock \bibinfo{journal}{Magyar Tudom\'anyos Akad\'emia Matematikai Kutat\'o
  Int\'ezet K\"ozlem\'enyei} \bibinfo{volume}{8} (\bibinfo{year}{1963})
  \bibinfo{pages}{229--243}.
\bibitem[{Doerr et~al.(2016)Doerr, Doerr, and Yang}]{DoerrDY16}
\bibinfo{author}{B.~Doerr}, \bibinfo{author}{C.~Doerr},
  \bibinfo{author}{J.~Yang},
\newblock \bibinfo{title}{Optimal parameter choices via precise black-box
  analysis},
\newblock in: \bibinfo{booktitle}{Proc. of Genetic and Evolutionary Computation
  Conference (GECCO'16)}, \bibinfo{publisher}{{ACM}}, \bibinfo{year}{2016}, pp.
  \bibinfo{pages}{1123--1130}.
\bibitem[{Doerr et~al.(2018)Doerr, Ye, van Rijn, Wang, and
  B{\"{a}}ck}]{DoerrYRWB18}
\bibinfo{author}{C.~Doerr}, \bibinfo{author}{F.~Ye}, \bibinfo{author}{S.~van
  Rijn}, \bibinfo{author}{H.~Wang}, \bibinfo{author}{T.~B{\"{a}}ck},
\newblock \bibinfo{title}{Towards a theory-guided benchmarking suite for
  discrete black-box optimization heuristics: profiling {(1} + {\(\lambda\)})
  {EA} variants on {O}ne{M}ax and {L}eading{O}nes},
\newblock in: \bibinfo{booktitle}{Proc. of Genetic and Evolutionary Computation
  Conference (GECCO'18)}, \bibinfo{publisher}{ACM}, \bibinfo{year}{2018}, pp.
  \bibinfo{pages}{951--958}. \URLprefix
  \url{https://doi.org/10.1145/3205455.3205621}.
  \DOIprefix\doi{10.1145/3205455.3205621}.
\bibitem[{Doerr(2018)}]{Doerr18BBC}
\bibinfo{author}{C.~Doerr},
\newblock \bibinfo{title}{Complexity theory for discrete black-box optimization
  heuristics},
\newblock \bibinfo{journal}{CoRR} \bibinfo{volume}{abs/1801.02037}
  (\bibinfo{year}{2018}). \href{http://arxiv.org/abs/1801.02037}{{\tt
  arXiv:1801.02037}}, \bibinfo{note}{available at
  \url{http://arxiv.org/abs/1801.02037}}.
\bibitem[{Thierens(2009)}]{Thierens09}
\bibinfo{author}{D.~Thierens},
\newblock \bibinfo{title}{On benchmark properties for adaptive operator
  selection},
\newblock in: \bibinfo{booktitle}{Proc. of Genetic and Evolutionary Computation
  Conference (GECCO'09)}, \bibinfo{publisher}{ACM}, \bibinfo{year}{2009}, pp.
  \bibinfo{pages}{2217--2218}. \URLprefix
  \url{https://doi.org/10.1145/1570256.1570306}.
  \DOIprefix\doi{10.1145/1570256.1570306}.
\bibitem[{Doerr(2018)}]{Doerr18domi}
\bibinfo{author}{B.~Doerr},
\newblock \bibinfo{title}{Better runtime guarantees via stochastic domination},
\newblock in: \bibinfo{booktitle}{Proc. of Evolutionary Computation in
  Combinatorial Optimization (EvoCOP'18)}, volume \bibinfo{volume}{10782} of
  \textit{\bibinfo{series}{Lecture Notes in Computer Science}},
  \bibinfo{publisher}{Springer}, \bibinfo{year}{2018}, pp.
  \bibinfo{pages}{1--17}. \URLprefix
  \url{https://doi.org/10.1007/978-3-319-77449-7\_1}.
  \DOIprefix\doi{10.1007/978-3-319-77449-7\_1}, \bibinfo{note}{full version
  available at \url{http://arxiv.org/abs/1801.04487}}.
\bibitem[{Doerr and Lengler(2018)}]{DoerrL17LO}
\bibinfo{author}{C.~Doerr}, \bibinfo{author}{J.~Lengler},
\newblock \bibinfo{title}{The (1+1) elitist black-box complexity of
  {L}eading{O}nes},
\newblock \bibinfo{journal}{Algorithmica} \bibinfo{volume}{80}
  (\bibinfo{year}{2018}) \bibinfo{pages}{1579--1603}. \URLprefix
  \url{https://doi.org/10.1007/s00453-017-0304-6}.
\bibitem[{Afshani et~al.(2019)Afshani, Agrawal, Doerr, Doerr, Larsen, and
  Mehlhorn}]{AfshaniADLMW19}
\bibinfo{author}{P.~Afshani}, \bibinfo{author}{M.~Agrawal},
  \bibinfo{author}{B.~Doerr}, \bibinfo{author}{C.~Doerr},
  \bibinfo{author}{K.~G. Larsen}, \bibinfo{author}{K.~Mehlhorn},
\newblock \bibinfo{title}{The query complexity of a permutation-based variant
  of mastermind},
\newblock \bibinfo{journal}{Discrete Applied Mathematics}
  (\bibinfo{year}{2019}). \DOIprefix\doi{10.1016/j.dam.2019.01.007},
  \bibinfo{note}{in press}.
\bibitem[{Doerr and Winzen(2012)}]{DoerrW11EA}
\bibinfo{author}{B.~Doerr}, \bibinfo{author}{C.~Winzen},
\newblock \bibinfo{title}{Black-box complexity: Breaking the {$O(n \log n)$}
  barrier of {L}eading{O}nes},
\newblock in: \bibinfo{booktitle}{Artificial Evolution (EA'11), Revised
  Selected Papers}, volume \bibinfo{volume}{7401} of
  \textit{\bibinfo{series}{Lecture Notes in Computer Science}},
  \bibinfo{publisher}{Springer}, \bibinfo{year}{2012}, pp.
  \bibinfo{pages}{205--216}.
\bibitem[{Doerr et~al.(2011)Doerr, Johannsen, K{\"o}tzing, Lehre, Wagner, and
  Winzen}]{DoerrJKLWW11}
\bibinfo{author}{B.~Doerr}, \bibinfo{author}{D.~Johannsen},
  \bibinfo{author}{T.~K{\"o}tzing}, \bibinfo{author}{P.~K. Lehre},
  \bibinfo{author}{M.~Wagner}, \bibinfo{author}{C.~Winzen},
\newblock \bibinfo{title}{Faster black-box algorithms through higher arity
  operators},
\newblock in: \bibinfo{booktitle}{Proc. of Foundations of Genetic Algorithms
  (FOGA'11)}, \bibinfo{publisher}{ACM}, \bibinfo{year}{2011}, pp.
  \bibinfo{pages}{163--172}.
\bibitem[{Droste et~al.(2002)Droste, Jansen, and Wegener}]{DrosteJW02}
\bibinfo{author}{S.~Droste}, \bibinfo{author}{T.~Jansen},
  \bibinfo{author}{I.~Wegener},
\newblock \bibinfo{title}{On the analysis of the (1+1) evolutionary algorithm},
\newblock \bibinfo{journal}{Theoretical Computer Science} \bibinfo{volume}{276}
  (\bibinfo{year}{2002}) \bibinfo{pages}{51--81}.
\bibitem[{Witt(2013)}]{Witt13j}
\bibinfo{author}{C.~Witt},
\newblock \bibinfo{title}{Tight bounds on the optimization time of a randomized
  search heuristic on linear functions},
\newblock \bibinfo{journal}{Combinatorics, Probability {\&} Computing}
  \bibinfo{volume}{22} (\bibinfo{year}{2013}) \bibinfo{pages}{294--318}.
\bibitem[{Weise and Wu(2018)}]{Wmodel}
\bibinfo{author}{T.~Weise}, \bibinfo{author}{Z.~Wu},
\newblock \bibinfo{title}{Difficult features of combinatorial optimization
  problems and the tunable w-model benchmark problem for simulating them},
\newblock in: \bibinfo{booktitle}{Proc. of Genetic and Evolutionary Computation
  Conference (GECCO'18), Companion Material)}, \bibinfo{publisher}{ACM},
  \bibinfo{year}{2018}, pp. \bibinfo{pages}{1769--1776}.
  \DOIprefix\doi{10.1145/3205651.3208240}.
\bibitem[{Weise(2018)}]{Wmodelpractice}
\bibinfo{author}{T.~Weise}, \bibinfo{title}{The {W-Model}, a tunable black-box
  discrete optimization benchmarking ({BB-DOB}) problem, implemented for the
  {BB-DOB@GECCO} workshop}, \bibinfo{year}{2018}. \bibinfo{note}{Data is
  available at \url{https://github.com/thomasWeise/BBDOB_W_Model}}.
\bibitem[{Kauffman and Levin(1987)}]{NKlandscapes}
\bibinfo{author}{S.~Kauffman}, \bibinfo{author}{S.~Levin},
\newblock \bibinfo{title}{Towards a general theory of adaptive walks on rugged
  landscapes},
\newblock \bibinfo{journal}{Journal of Theoretical Biology}
  \bibinfo{volume}{128} (\bibinfo{year}{1987}) \bibinfo{pages}{11--45}.
\bibitem[{Einarsson et~al.(2018)Einarsson, Lengler, Gauy, Meier, Mujika,
  Steger, and Weissenberger}]{EinarssonLGMMSW18}
\bibinfo{author}{H.~Einarsson}, \bibinfo{author}{J.~Lengler},
  \bibinfo{author}{M.~M. Gauy}, \bibinfo{author}{F.~Meier},
  \bibinfo{author}{A.~Mujika}, \bibinfo{author}{A.~Steger},
  \bibinfo{author}{F.~Weissenberger},
\newblock \bibinfo{title}{The linear hidden subset problem for the {(1} + 1)
  {EA} with scheduled and adaptive mutation rates},
\newblock in: \bibinfo{booktitle}{Proc. of Genetic and Evolutionary Computation
  Conference (GECCO'18)}, \bibinfo{publisher}{ACM}, \bibinfo{year}{2018}, pp.
  \bibinfo{pages}{1491--1498}. \URLprefix
  \url{https://doi.org/10.1145/3205455.3205519}.
  \DOIprefix\doi{10.1145/3205455.3205519}.
\bibitem[{Doerr et~al.(2017)Doerr, Doerr, and K{\"{o}}tzing}]{DoerrDK17}
\bibinfo{author}{B.~Doerr}, \bibinfo{author}{C.~Doerr},
  \bibinfo{author}{T.~K{\"{o}}tzing},
\newblock \bibinfo{title}{Unknown solution length problems with no
  asymptotically optimal run time},
\newblock in: \bibinfo{booktitle}{Proc. of Genetic and Evolutionary Computation
  Conference (GECCO'17)}, \bibinfo{publisher}{ACM}, \bibinfo{year}{2017}, pp.
  \bibinfo{pages}{1367--1374}. \URLprefix
  \url{http://doi.acm.org/10.1145/3071178.3071233}.
\bibitem[{Shapiro et~al.(1968)Shapiro, Pettengill, Ash, Stone, Smith, Ingalls,
  and Brockelman}]{ShapiroPRL1968}
\bibinfo{author}{I.~I. Shapiro}, \bibinfo{author}{G.~H. Pettengill},
  \bibinfo{author}{M.~E. Ash}, \bibinfo{author}{M.~L. Stone},
  \bibinfo{author}{W.~B. Smith}, \bibinfo{author}{R.~P. Ingalls},
  \bibinfo{author}{R.~A. Brockelman},
\newblock \bibinfo{title}{Fourth test of general relativity: Preliminary
  results},
\newblock \bibinfo{journal}{Phys. Rev. Lett.} \bibinfo{volume}{20}
  (\bibinfo{year}{1968}) \bibinfo{pages}{1265--1269}.
  \DOIprefix\doi{10.1103/PhysRevLett.20.1265}.
\bibitem[{Pasha et~al.(2000)Pasha, Moharir, and Rao}]{Pasha2000}
\bibinfo{author}{I.~A. Pasha}, \bibinfo{author}{P.~S. Moharir},
  \bibinfo{author}{N.~S. Rao},
\newblock \bibinfo{title}{Bi-alphabetic pulse compression radar signal design},
\newblock \bibinfo{journal}{Sadhana} \bibinfo{volume}{25}
  (\bibinfo{year}{2000}) \bibinfo{pages}{481--488}.
  \DOIprefix\doi{10.1007/BF02703629}.
\bibitem[{Militzer et~al.(1998)Militzer, Zamparelli, and Beule}]{LAC_Militzer}
\bibinfo{author}{B.~Militzer}, \bibinfo{author}{M.~Zamparelli},
  \bibinfo{author}{D.~Beule},
\newblock \bibinfo{title}{Evolutionary search for low autocorrelated binary
  sequences},
\newblock \bibinfo{journal}{IEEE Transactions on Evolutionary Computation}
  \bibinfo{volume}{2} (\bibinfo{year}{1998}) \bibinfo{pages}{34--39}.
  \DOIprefix\doi{10.1109/4235.728212}.
\bibitem[{Packebusch and Mertens(2016)}]{LABS_Packebusch2016}
\bibinfo{author}{T.~Packebusch}, \bibinfo{author}{S.~Mertens},
\newblock \bibinfo{title}{Low autocorrelation binary sequences},
\newblock \bibinfo{journal}{Journal of Physics A: Mathematical and Theoretical}
  \bibinfo{volume}{49} (\bibinfo{year}{2016}) \bibinfo{pages}{165001}.
\bibitem[{{Barahona}(1982)}]{Ising_Barahona1982}
\bibinfo{author}{F.~{Barahona}},
\newblock \bibinfo{title}{{On the computational complexity of {Ising} spin
  glass models}},
\newblock \bibinfo{journal}{Journal of Physics A Mathematical General}
  \bibinfo{volume}{15} (\bibinfo{year}{1982}) \bibinfo{pages}{3241--3253}.
  \DOIprefix\doi{10.1088/0305-4470/15/10/028}.
\bibitem[{{Lucas}(2014)}]{Ising_Lucas2014}
\bibinfo{author}{A.~{Lucas}},
\newblock \bibinfo{title}{{Ising formulations of many NP problems}},
\newblock \bibinfo{journal}{Frontiers in Physics} \bibinfo{volume}{2}
  (\bibinfo{year}{2014}) \bibinfo{pages}{5}.
  \DOIprefix\doi{10.3389/fphy.2014.00005}.
  \href{http://arxiv.org/abs/1302.5843}{{\tt arXiv:1302.5843}}.
\bibitem[{Briest et~al.(2004)Briest, Brockhoff, Degener, Englert, Gunia,
  Heering, Jansen, Leifhelm, Plociennik, R{\"{o}}glin, Schweer, Sudholt,
  Tannenbaum, and Wegener}]{Briest04}
\bibinfo{author}{P.~Briest}, \bibinfo{author}{D.~Brockhoff},
  \bibinfo{author}{B.~Degener}, \bibinfo{author}{M.~Englert},
  \bibinfo{author}{C.~Gunia}, \bibinfo{author}{O.~Heering},
  \bibinfo{author}{T.~Jansen}, \bibinfo{author}{M.~Leifhelm},
  \bibinfo{author}{K.~Plociennik}, \bibinfo{author}{H.~R{\"{o}}glin},
  \bibinfo{author}{A.~Schweer}, \bibinfo{author}{D.~Sudholt},
  \bibinfo{author}{S.~Tannenbaum}, \bibinfo{author}{I.~Wegener},
\newblock \bibinfo{title}{The {Ising} model: Simple evolutionary algorithms as
  adaptation schemes},
\newblock in: \bibinfo{booktitle}{Parallel Problem Solving from Nature - {PPSN}
  VIII, 8th International Conference, Birmingham, UK, September 18-22, 2004,
  Proceedings}, \bibinfo{year}{2004}, pp. \bibinfo{pages}{31--40}. \URLprefix
  \url{https://doi.org/10.1007/978-3-540-30217-9\_4}.
  \DOIprefix\doi{10.1007/978-3-540-30217-9\_4}.
\bibitem[{Fischer and Wegener(2005)}]{FischerWegener2005Ising}
\bibinfo{author}{S.~Fischer}, \bibinfo{author}{I.~Wegener},
\newblock \bibinfo{title}{The one-dimensional {I}sing model: Mutation versus
  recombination},
\newblock \bibinfo{journal}{Theoretical Computer Science} \bibinfo{volume}{344}
  (\bibinfo{year}{2005}) \bibinfo{pages}{208--225}.
\bibitem[{Sudholt(2005)}]{Sudholt2005Crossover}
\bibinfo{author}{D.~Sudholt},
\newblock \bibinfo{title}{Crossover is provably essential for the {I}sing model
  on trees},
\newblock in: \bibinfo{booktitle}{Proc. of Genetic and Evolutionary Computation
  Conference (GECCO'05)}, \bibinfo{year}{2005}, pp.
  \bibinfo{pages}{1161--1167}.
\bibitem[{Mellor(2011)}]{Mellor_thesis}
\bibinfo{author}{V.~Mellor},
\newblock \bibinfo{title}{Numerical simulations of the {I}sing model on the
  union jack lattice},
\newblock \bibinfo{journal}{{arXiv}} \bibinfo{volume}{1101.5015}
  (\bibinfo{year}{2011}). \bibinfo{note}{Available at
  \url{https://arxiv.org/abs/1101.5015}}.
\bibitem[{B{\"a}ck and Khuri(1994)}]{BK94}
\bibinfo{author}{T.~B{\"a}ck}, \bibinfo{author}{S.~Khuri},
\newblock \bibinfo{title}{An evolutionary heuristic for the maximum independent
  set problem},
\newblock in: \bibinfo{booktitle}{Proc.~1st IEEE Conference on Evolutionary
  Computation}, \bibinfo{publisher}{IEEE}, \bibinfo{year}{1994}, pp.
  \bibinfo{pages}{531--535}. \DOIprefix\doi{10.1109/ICEC.1994.350004}.
\bibitem[{Bell and Stevens(2009)}]{NQP_Bell2009}
\bibinfo{author}{J.~Bell}, \bibinfo{author}{B.~Stevens},
\newblock \bibinfo{title}{A survey of known results and research areas for
  {N-queens}},
\newblock \bibinfo{journal}{Discrete Math.} \bibinfo{volume}{309}
  (\bibinfo{year}{2009}) \bibinfo{pages}{1--31}. \URLprefix
  \url{http://dx.doi.org/10.1016/j.disc.2007.12.043}.
  \DOIprefix\doi{10.1016/j.disc.2007.12.043}.
\bibitem[{Rodionova et~al.(2019)Rodionova, Antonov, Buzdalova, and
  Doerr}]{Arina2019}
\bibinfo{author}{A.~Rodionova}, \bibinfo{author}{K.~Antonov},
  \bibinfo{author}{A.~Buzdalova}, \bibinfo{author}{C.~Doerr},
\newblock \bibinfo{title}{Offspring population size matters when comparing
  evolutionary algorithms with self-adjusting mutation rates},
\newblock in: \bibinfo{booktitle}{Proc. of Genetic and Evolutionary Computation
  Conference (GECCO'19)}, \bibinfo{publisher}{ACM}, \bibinfo{year}{2019}, pp.
  \bibinfo{pages}{855--863}. \URLprefix
  \url{https://doi.org/10.1145/3321707.3321827}.
  \DOIprefix\doi{10.1145/3321707.3321827}, \bibinfo{note}{full version
  available online at \url{https://arxiv.org/abs/1904.08032}}.
\bibitem[{Dang and Doerr(2019)}]{DangD19}
\bibinfo{author}{N.~Dang}, \bibinfo{author}{C.~Doerr},
\newblock \bibinfo{title}{Hyper-parameter tuning for the {(1} +
  (\emph{{\(\lambda\)}, {\(\lambda\)}})) {GA}},
\newblock in: \bibinfo{booktitle}{Proc. of Genetic and Evolutionary Computation
  Conference (GECCO'19)}, \bibinfo{publisher}{ACM}, \bibinfo{year}{2019}, pp.
  \bibinfo{pages}{889--897}. \URLprefix
  \url{https://doi.org/10.1145/3321707.3321725}.
  \DOIprefix\doi{10.1145/3321707.3321725}.
\bibitem[{{Carvalho Pinto} and Doerr(2017)}]{CarvalhoD17}
\bibinfo{author}{E.~{Carvalho Pinto}}, \bibinfo{author}{C.~Doerr},
\newblock \bibinfo{title}{Discussion of a more practice-aware runtime analysis
  for evolutionary algorithms},
\newblock in: \bibinfo{booktitle}{Proc. of Artificial Evolution (EA'17)},
  \bibinfo{year}{2017}, pp. \bibinfo{pages}{298--305}. \URLprefix
  \url{https://ea2017.inria.fr//EA2017_Proceedings_web_ISBN_978-2-9539267-7-4.pdf},
  \bibinfo{note}{full version available at
  \url{http://arxiv.org/abs/1812.00493}}.
\bibitem[{Doerr et~al.(2017{\natexlab{a}})Doerr, Le, Makhmara, and
  Nguyen}]{FastGA17}
\bibinfo{author}{B.~Doerr}, \bibinfo{author}{H.~P. Le},
  \bibinfo{author}{R.~Makhmara}, \bibinfo{author}{T.~D. Nguyen},
\newblock \bibinfo{title}{Fast genetic algorithms},
\newblock in: \bibinfo{booktitle}{Proc. of Genetic and Evolutionary Computation
  Conference (GECCO'17)}, \bibinfo{publisher}{ACM},
  \bibinfo{year}{2017}{\natexlab{a}}, pp. \bibinfo{pages}{777--784}. \URLprefix
  \url{http://doi.acm.org/10.1145/3071178.3071301}.
  \DOIprefix\doi{10.1145/3071178.3071301}.
\bibitem[{Doerr et~al.(2017{\natexlab{b}})Doerr, Gie{\ss}en, Witt, and
  Yang}]{DoerrGWY17}
\bibinfo{author}{B.~Doerr}, \bibinfo{author}{C.~Gie{\ss}en},
  \bibinfo{author}{C.~Witt}, \bibinfo{author}{J.~Yang},
\newblock \bibinfo{title}{The $(1+\lambda)$~evolutionary algorithm with
  self-adjusting mutation rate},
\newblock in: \bibinfo{booktitle}{Proc. of Genetic and Evolutionary Computation
  Conference (GECCO'17)}, \bibinfo{publisher}{ACM},
  \bibinfo{year}{2017}{\natexlab{b}}, pp. \bibinfo{pages}{1351--1358}.
\bibitem[{{Ye} et~al.(2019){Ye}, {Doerr}, and {B{\"a}ck}}]{YeDB19}
\bibinfo{author}{F.~{Ye}}, \bibinfo{author}{C.~{Doerr}},
  \bibinfo{author}{T.~{B{\"a}ck}},
\newblock \bibinfo{title}{{Interpolating Local and Global Search by Controlling
  the Variance of Standard Bit Mutation}},
\newblock in: \bibinfo{booktitle}{Proc. of Congress on Evolutionary Computation
  (CEC'19)}, \bibinfo{publisher}{IEEE}, \bibinfo{year}{2019}, pp.
  \bibinfo{pages}{2292--2299}. \bibinfo{note}{Also available at
  \url{http://arxiv.org/abs/1901.05573}}.
\bibitem[{B{\"{a}}ck and Sch{\"{u}}tz(1996)}]{BackS96}
\bibinfo{author}{T.~B{\"{a}}ck}, \bibinfo{author}{M.~Sch{\"{u}}tz},
\newblock \bibinfo{title}{Intelligent mutation rate control in canonical
  genetic algorithms},
\newblock in: \bibinfo{booktitle}{International Symposium on Foundations of
  Intelligent Systems (ISMIS'96)}, volume \bibinfo{volume}{1079} of
  \textit{\bibinfo{series}{Lecture Notes in Computer Science}},
  \bibinfo{publisher}{Springer}, \bibinfo{year}{1996}, pp.
  \bibinfo{pages}{158--167}.
\bibitem[{Goldberg(1989)}]{Goldberg}
\bibinfo{author}{D.~Goldberg}, \bibinfo{title}{{Genetic Algorithms in Search,
  Optimization, and Machine Learning}}, \bibinfo{publisher}{Addison Wesley},
  \bibinfo{address}{Reading, MA}, \bibinfo{year}{1989}.
\bibitem[{B{\"a}ck(1996)}]{Baeck-book}
\bibinfo{author}{T.~B{\"a}ck}, \bibinfo{title}{{E}volutionary {A}lgorithms in
  {T}heory and {P}ractice}, \bibinfo{publisher}{Oxford University Press},
  \bibinfo{address}{New York, NY, USA}, \bibinfo{year}{1996}.
\bibitem[{M\"uhlenbein(1997)}]{Muhlenbein97}
\bibinfo{author}{H.~M\"uhlenbein},
\newblock \bibinfo{title}{The equation for response to selection and its use
  for prediction},
\newblock \bibinfo{journal}{Evolutionary Computation} \bibinfo{volume}{5}
  (\bibinfo{year}{1997}) \bibinfo{pages}{303--346}. \URLprefix
  \url{https://doi.org/10.1162/evco.1997.5.3.303}.
  \DOIprefix\doi{10.1162/evco.1997.5.3.303}.
\bibitem[{M{\"u}hlenbein and Paa{\ss}(1996)}]{MuhlenbeinV96}
\bibinfo{author}{H.~M{\"u}hlenbein}, \bibinfo{author}{G.~Paa{\ss}},
\newblock \bibinfo{title}{From recombination of genes to the estimation of
  distributions i. binary parameters},
\newblock in: \bibinfo{editor}{H.-M. Voigt}, \bibinfo{editor}{W.~Ebeling},
  \bibinfo{editor}{I.~Rechenberg}, \bibinfo{editor}{H.-P. Schwefel} (Eds.),
  \bibinfo{booktitle}{Proc. of Parallel Problem Solving from Nature (PPSN'96)},
  \bibinfo{publisher}{Springer}, \bibinfo{year}{1996}, pp.
  \bibinfo{pages}{178--187}.
\bibitem[{Krejca and Witt(2018)}]{KrejcaW18}
\bibinfo{author}{M.~S. Krejca}, \bibinfo{author}{C.~Witt},
\newblock \bibinfo{title}{Theory of estimation-of-distribution algorithms},
\newblock \bibinfo{journal}{CoRR} \bibinfo{volume}{abs/1806.05392}
  (\bibinfo{year}{2018}). \URLprefix \url{http://arxiv.org/abs/1806.05392}.
\bibitem[{Horesh et~al.(2019)Horesh, B{\"{a}}ck, and Shir}]{HoreshBS19}
\bibinfo{author}{N.~Horesh}, \bibinfo{author}{T.~B{\"{a}}ck},
  \bibinfo{author}{O.~M. Shir},
\newblock \bibinfo{title}{Predict or screen your expensive assay: Doe vs.
  surrogates in experimental combinatorial optimization},
\newblock in: \bibinfo{booktitle}{Proc. of Genetic and Evolutionary Computation
  Conference (GECCO'19)}, \bibinfo{publisher}{ACM}, \bibinfo{year}{2019}, pp.
  \bibinfo{pages}{274--284}. \URLprefix
  \url{https://doi.org/10.1145/3321707.3321801}.
  \DOIprefix\doi{10.1145/3321707.3321801}.
\bibitem[{Hansen(2018)}]{Hansen18}
\bibinfo{author}{N.~Hansen},
\newblock \bibinfo{title}{A practical guide to experimentation},
\newblock in: \bibinfo{booktitle}{Proc. of Genetic and Evolutionary Computation
  Conference (GECCO'18), Companion material}, \bibinfo{publisher}{ACM},
  \bibinfo{year}{2018}, pp. \bibinfo{pages}{432--447}.
\bibitem[{Calvo et~al.(2019)Calvo, Shir, Ceberio, Doerr, Wang, B{\"{a}}ck, and
  Lozano}]{CalvoSCD0BL19}
\bibinfo{author}{B.~Calvo}, \bibinfo{author}{O.~M. Shir},
  \bibinfo{author}{J.~Ceberio}, \bibinfo{author}{C.~Doerr},
  \bibinfo{author}{H.~Wang}, \bibinfo{author}{T.~B{\"{a}}ck},
  \bibinfo{author}{J.~A. Lozano},
\newblock \bibinfo{title}{Bayesian performance analysis for black-box
  optimization benchmarking},
\newblock in: \bibinfo{booktitle}{Proc. of the Genetic and Evolutionary
  Computation Conference (GECCO'19, Companion Material)},
  \bibinfo{publisher}{ACM}, \bibinfo{year}{2019}, pp.
  \bibinfo{pages}{1789--1797}. \URLprefix
  \url{https://doi.org/10.1145/3319619.3326888}.
  \DOIprefix\doi{10.1145/3319619.3326888}.
\bibitem[{Calvo et~al.(2018)Calvo, Ceberio, and Lozano}]{CalvoCL18}
\bibinfo{author}{B.~Calvo}, \bibinfo{author}{J.~Ceberio},
  \bibinfo{author}{J.~A. Lozano},
\newblock \bibinfo{title}{Bayesian inference for algorithm ranking analysis},
\newblock in: \bibinfo{booktitle}{Proc. of Genetic and Evolutionary Computation
  Conference (GECCO'18), companion material}, \bibinfo{publisher}{ACM},
  \bibinfo{year}{2018}, pp. \bibinfo{pages}{324--325}. \URLprefix
  \url{https://doi.org/10.1145/3205651.3205658}.
  \DOIprefix\doi{10.1145/3205651.3205658}.
\bibitem[{Hansen et~al.(2016)Hansen, Auger, Brockhoff, Tusar, and
  Tu\v{s}ar}]{HansenABTT16}
\bibinfo{author}{N.~Hansen}, \bibinfo{author}{A.~Auger},
  \bibinfo{author}{D.~Brockhoff}, \bibinfo{author}{D.~Tusar},
  \bibinfo{author}{T.~Tu\v{s}ar},
\newblock \bibinfo{title}{{COCO:} performance assessment},
\newblock \bibinfo{journal}{CoRR} \bibinfo{volume}{abs/1605.03560}
  (\bibinfo{year}{2016}). \URLprefix \url{http://arxiv.org/abs/1605.03560}.
\bibitem[{Hastie et~al.(2013)Hastie, Tibshirani, and
  Friedman}]{hastie2013elements}
\bibinfo{author}{T.~Hastie}, \bibinfo{author}{R.~Tibshirani},
  \bibinfo{author}{J.~Friedman}, \bibinfo{title}{The Elements of Statistical
  Learning: Data Mining, Inference, and Prediction}, Springer Series in
  Statistics, \bibinfo{publisher}{Springer New York}, \bibinfo{year}{2013}.
\bibitem[{Fonseca et~al.(2011)Fonseca, Guerreiro,
  L{\'{o}}pez{-}Ib{\'{a}}{\~{n}}ez, and Paquete}]{eaf}
\bibinfo{author}{C.~M. Fonseca}, \bibinfo{author}{A.~P. Guerreiro},
  \bibinfo{author}{M.~L{\'{o}}pez{-}Ib{\'{a}}{\~{n}}ez},
  \bibinfo{author}{L.~Paquete},
\newblock \bibinfo{title}{On the computation of the empirical attainment
  function},
\newblock in: \bibinfo{booktitle}{Proc. of Evolutionary Multi-Criterion
  Optimization (EMO'11)}, volume \bibinfo{volume}{6576} of
  \textit{\bibinfo{series}{Lecture Notes in Computer Science}},
  \bibinfo{publisher}{Springer}, \bibinfo{year}{2011}, pp.
  \bibinfo{pages}{106--120}. \URLprefix
  \url{https://doi.org/10.1007/978-3-642-19893-9\_8}.
  \DOIprefix\doi{10.1007/978-3-642-19893-9\_8}.
\bibitem[{Mor{\'e} and Wild(2009)}]{more2009benchmarking}
\bibinfo{author}{J.~J. Mor{\'e}}, \bibinfo{author}{S.~M. Wild},
\newblock \bibinfo{title}{Benchmarking derivative-free optimization
  algorithms},
\newblock \bibinfo{journal}{SIAM Journal on Optimization} \bibinfo{volume}{20}
  (\bibinfo{year}{2009}) \bibinfo{pages}{172--191}.
\bibitem[{Benavoli et~al.(2017)Benavoli, Corani, Demsar, and
  Zaffalon}]{BenavoliCDZ17}
\bibinfo{author}{A.~Benavoli}, \bibinfo{author}{G.~Corani},
  \bibinfo{author}{J.~Demsar}, \bibinfo{author}{M.~Zaffalon},
\newblock \bibinfo{title}{Time for a change: a tutorial for comparing multiple
  classifiers through bayesian analysis},
\newblock \bibinfo{journal}{Journal of Machine Learning Research}
  \bibinfo{volume}{18} (\bibinfo{year}{2017}) \bibinfo{pages}{77:1--77:36}.
  \URLprefix \url{http://jmlr.org/papers/v18/16-305.html}.
\bibitem[{Belkhir et~al.(2017)Belkhir, Dr{\'{e}}o, Sav{\'{e}}ant, and
  Schoenauer}]{BelkhirDSS17}
\bibinfo{author}{N.~Belkhir}, \bibinfo{author}{J.~Dr{\'{e}}o},
  \bibinfo{author}{P.~Sav{\'{e}}ant}, \bibinfo{author}{M.~Schoenauer},
\newblock \bibinfo{title}{Per instance algorithm configuration of {CMA-ES} with
  limited budget},
\newblock in: \bibinfo{booktitle}{Proc. of Genetic and Evolutionary Computation
  Conference (GECCO'17)}, \bibinfo{publisher}{ACM}, \bibinfo{year}{2017}, pp.
  \bibinfo{pages}{681--688}. \URLprefix
  \url{https://doi.org/10.1145/3071178.3071343}.
  \DOIprefix\doi{10.1145/3071178.3071343}.
\bibitem[{Kerschke and Trautmann(2016)}]{flacco}
\bibinfo{author}{P.~Kerschke}, \bibinfo{author}{H.~Trautmann},
\newblock \bibinfo{title}{The r-package {FLACCO} for exploratory landscape
  analysis with applications to multi-objective optimization problems},
\newblock in: \bibinfo{booktitle}{Proc. of Congress on Evolutionary Computation
  (CEC'16)}, \bibinfo{publisher}{IEEE}, \bibinfo{year}{2016}, pp.
  \bibinfo{pages}{5262--5269}. \URLprefix
  \url{https://doi.org/10.1109/CEC.2016.7748359}.
  \DOIprefix\doi{10.1109/CEC.2016.7748359}.
\bibitem[{Thomson et~al.(2018{\natexlab{a}})Thomson, V{\'{e}}rel, Ochoa,
  Veerapen, and McMenemy}]{ThomsonVOVM18}
\bibinfo{author}{S.~L. Thomson}, \bibinfo{author}{S.~V{\'{e}}rel},
  \bibinfo{author}{G.~Ochoa}, \bibinfo{author}{N.~Veerapen},
  \bibinfo{author}{P.~McMenemy},
\newblock \bibinfo{title}{On the fractal nature of local optima networks},
\newblock in: \bibinfo{booktitle}{Proc. of Evolutionary Computation in
  Combinatorial Optimization (EvoCOP'18)}, volume \bibinfo{volume}{10782} of
  \textit{\bibinfo{series}{Lecture Notes in Computer Science}},
  \bibinfo{publisher}{Springer}, \bibinfo{year}{2018}{\natexlab{a}}, pp.
  \bibinfo{pages}{18--33}. \URLprefix
  \url{https://doi.org/10.1007/978-3-319-77449-7\_2}.
  \DOIprefix\doi{10.1007/978-3-319-77449-7\_2}.
\bibitem[{Thomson et~al.(2018{\natexlab{b}})Thomson, V{\'{e}}rel, Ochoa,
  Veerapen, and Cairns}]{ThomsonVOVC18}
\bibinfo{author}{S.~L. Thomson}, \bibinfo{author}{S.~V{\'{e}}rel},
  \bibinfo{author}{G.~Ochoa}, \bibinfo{author}{N.~Veerapen},
  \bibinfo{author}{D.~Cairns},
\newblock \bibinfo{title}{Multifractality and dimensional determinism in local
  optima networks},
\newblock in: \bibinfo{booktitle}{Proc. of Genetic and Evolutionary Computation
  Conference ({GECCO}'18)}, \bibinfo{publisher}{ACM},
  \bibinfo{year}{2018}{\natexlab{b}}, pp. \bibinfo{pages}{371--378}. \URLprefix
  \url{https://doi.org/10.1145/3205455.3205472}.
  \DOIprefix\doi{10.1145/3205455.3205472}.
\bibitem[{Cahon et~al.(2004)Cahon, Melab, and Talbi}]{paradiseo}
\bibinfo{author}{S.~Cahon}, \bibinfo{author}{N.~Melab},
  \bibinfo{author}{E.~Talbi},
\newblock \bibinfo{title}{{ParadisEO}: {A} framework for the reusable design of
  parallel and distributed metaheuristics},
\newblock \bibinfo{journal}{J. Heuristics} \bibinfo{volume}{10}
  (\bibinfo{year}{2004}) \bibinfo{pages}{357--380}. \URLprefix
  \url{https://doi.org/10.1023/B:HEUR.0000026900.92269.ec}.
  \DOIprefix\doi{10.1023/B:HEUR.0000026900.92269.ec}.
\bibitem[{Durillo and Nebro(2011)}]{jMetal}
\bibinfo{author}{J.~J. Durillo}, \bibinfo{author}{A.~J. Nebro},
\newblock \bibinfo{title}{{jMetal}: A {Java} framework for multi-objective
  optimization},
\newblock \bibinfo{journal}{Advances in Engineering Software}
  \bibinfo{volume}{42} (\bibinfo{year}{2011}) \bibinfo{pages}{760--771}.
  \URLprefix
  \url{http://www.sciencedirect.com/science/article/pii/S0965997811001219}.
  \DOIprefix\doi{DOI: 10.1016/j.advengsoft.2011.05.014}.
\bibitem[{van Rijn et~al.(2016)van Rijn, Wang, van Leeuwen, and
  B{\"{a}}ck}]{modularCMAES}
\bibinfo{author}{S.~van Rijn}, \bibinfo{author}{H.~Wang},
  \bibinfo{author}{M.~van Leeuwen}, \bibinfo{author}{T.~B{\"{a}}ck},
\newblock \bibinfo{title}{Evolving the structure of evolution strategies},
\newblock in: \bibinfo{booktitle}{2016 {IEEE} Symposium Series on Computational
  Intelligence, {SSCI} 2016, Athens, Greece, December 6-9, 2016},
  \bibinfo{publisher}{IEEE}, \bibinfo{year}{2016}, pp. \bibinfo{pages}{1--8}.
  \URLprefix \url{https://doi.org/10.1109/SSCI.2016.7850138}.
  \DOIprefix\doi{10.1109/SSCI.2016.7850138}.

\end{thebibliography}

\end{document}